\documentclass{article}

\usepackage{microtype}
\usepackage{graphicx}
\usepackage{booktabs} %

\usepackage{hyperref}

\usepackage[accepted]{icml2021}

\usepackage{hyperref}
\usepackage{url}

\usepackage{amsmath}
\usepackage{amssymb}
\usepackage{subfig}
\usepackage{enumitem}
\usepackage{float}
\usepackage{dsfont}
\usepackage{multirow}
\usepackage{multicol}
\usepackage{stackengine}

\usepackage{tikz}

\usepackage{varioref}
\labelformat{equation}{(#1)}

\newcommand{\e}{\epsilon}
\newcommand{\y}{\gamma}
\newcommand{\al}{\alpha}

\newcommand{\g}{\nabla}
\newcommand{\E}{\mathbb{E}}
\newcommand{\N}{\mathcal{N}}
\newcommand{\lb}{\left [}
\newcommand{\rb}{\right ]}
\newcommand{\lp}{\left (}
\newcommand{\rp}{\right )}

\newcommand{\Loss}{\mathcal{L}}

\newtheorem{theorem}{Theorem}

\newtheorem{observation}{Observation}

\makeatletter
\let\ORG@Gscale@box\Gscale@box
\long\def\Gscale@box#1{%
  \xdef\thelastscalefactor{#1}%
  \ORG@Gscale@box{#1}}
\makeatother

\usepackage{bm}
\usepackage[outline]{contour}

\newcommand{\ctimes}[3]{
\begin{tikzpicture}
\draw[fill=white, draw=white] (0,0) circle (3pt);
\draw[color={rgb,255:red,#1;green,#2;blue,#3}, thick] (-0.3,0) -- (0.3,0);
\draw[fill=white,draw=white] (0,0) circle (2.5pt);
\draw[color={rgb,255:red,#1;green,#2;blue,#3}, ultra thick] (-0.08,-0.08) -- (0.08,0.08);
\draw[color={rgb,255:red,#1;green,#2;blue,#3}, ultra thick] (0.08,-0.08) -- (-0.08,0.08);
\end{tikzpicture}
\hspace{-3pt}
}

\newcommand{\cplus}[3]{
\begin{tikzpicture}
\draw[fill=white, draw=white] (0,0) circle (3pt);
\draw[color={rgb,255:red,#1;green,#2;blue,#3}, thick] (-0.3,0) -- (0.3,0);
\draw[fill=white,draw=white] (0,0) circle (3pt);
\draw[color={rgb,255:red,#1;green,#2;blue,#3}, ultra thick] (0,-0.1) -- (0,0.1);
\draw[color={rgb,255:red,#1;green,#2;blue,#3}, ultra thick] (0.1,0) -- (-0.1,0);
\end{tikzpicture}
\hspace{-3pt}
}

\newcommand{\ccirc}[3]{
\begin{tikzpicture}
   \draw[color={rgb,255:red,#1;green,#2;blue,#3}, thick] (-0.3,0) -- (0.3,0);
   \draw[fill={rgb,255:red,#1;green,#2;blue,#3}, draw=white] (0,0) circle (3pt);
\end{tikzpicture}
\hspace{-3pt}
}

\newcommand{\csquare}[3]{
\begin{tikzpicture}
   \draw[color={rgb,255:red,#1;green,#2;blue,#3}, thick] (-0.3,0) -- (0.3,0);
   \draw[fill={rgb,255:red,#1;green,#2;blue,#3}, draw=white] (-0.1,-0.1) rectangle (0.1,0.1);
\end{tikzpicture}
\hspace{-3pt}
}

\newcommand{\ctriangle}[3]{
\begin{tikzpicture}
   \draw[color={rgb,255:red,#1;green,#2;blue,#3}, thick] (-0.3,0) -- (0.3,0);
   \draw[fill={rgb,255:red,#1;green,#2;blue,#3}, draw=white] (0.1,0.1) -- (-0.1,0.1) -- (0,-0.1) -- cycle;
\end{tikzpicture}
\hspace{-3pt}
}

\newcommand{\cdashed}{
\begin{tikzpicture}   
\draw[fill=white, draw=white] (0,0) circle (3pt);
\draw[color=black, ultra thick, densely dashed] (-0.2,0) -- (0.2,0);
\end{tikzpicture}
}

\icmltitlerunning{Marginalized Importance Sampling with the Successor Representation}

\begin{document}

\twocolumn[
\icmltitle{A Deep Reinforcement Learning Approach to Marginalized Importance Sampling with the Successor Representation}

\icmlsetsymbol{equal}{*}

\begin{icmlauthorlist}
\icmlauthor{Scott Fujimoto}{mcgill}
\icmlauthor{David Meger}{mcgill}
\icmlauthor{Doina Precup}{mcgill}
\end{icmlauthorlist}

\icmlaffiliation{mcgill}{Mila, McGill University}

\icmlcorrespondingauthor{Scott Fujimoto}{scott.fujimoto@mail.mcgill.ca}

\icmlkeywords{Reinforcement Learning, Marginalized Importance Sampling, DICE, Off-policy evaluation, Machine Learning, ICML}

\vskip 0.3in
]

\printAffiliationsAndNotice{}  %

\begin{abstract}
Marginalized importance sampling (MIS), which measures the density ratio between the state-action occupancy of a target policy and that of a sampling distribution, is a promising approach for off-policy evaluation. However, current state-of-the-art MIS methods rely on complex optimization tricks and succeed mostly on simple toy problems. We bridge the gap between MIS and deep reinforcement learning by observing that the density ratio can be computed from the successor representation of the target policy. The successor representation can be trained through deep reinforcement learning methodology and decouples the reward optimization from the dynamics of the environment, making the resulting algorithm stable and applicable to high-dimensional domains. We evaluate the empirical performance of our approach on a variety of challenging Atari and MuJoCo environments.
 \end{abstract}

\section{Introduction}

Off-policy evaluation (OPE) is a reinforcement learning (RL) task where the aim is to measure the performance of a target policy from data collected by a separate behavior policy \cite{sutton1998reinforcement}. As it can often be difficult or costly to obtain new data, OPE offers an avenue for re-using previously gathered data, making OPE an important challenge for applying RL to real-world domains~\cite{zhao2009reinforcement,mandel2014offline,swaminathan2017off,gauci2018horizon}. 

Marginalized importance sampling (MIS) \citep{liu2018breaking, xie2019towards, nachum2019dualdice} is a family of OPE methods which re-weight sampled rewards by directly learning the density ratio between the state-action occupancy of the target policy and the sampling distribution.
This approach can have significantly lower variance than traditional importance sampling methods~\citep{precup2001off}, which consider a product of ratios over trajectories, and is amenable to deterministic policies and behavior agnostic settings where the sampling distribution is unknown. 
However, the body of MIS work is largely theoretical, 
and as a result, empirical evaluations of MIS have mostly been carried out on simple low-dimensional tasks, such as mountain car~(state dim.\ of $2$) or cartpole~(state dim.\ of $4$). In comparison, deep RL algorithms have shown successful behaviors in high-dimensional domains such as Humanoid locomotion~(state dim.\ of $376$) and Atari (image-based). 

In this paper, we present a straightforward approach for MIS that can be computed from the successor representation~(SR)~\citep{dayan1993improving} of the target policy by directly optimizing the reward function. Our algorithm, the Successor Representation DIstribution Correction Estimation (SR-DICE), is the first method that allows MIS to scale to high-dimensional systems, far outperforming previous approaches.
In comparison to previous algorithms which rely on minimax optimization or kernel methods~\citep{liu2018breaking, nachum2019dualdice, uehara2019minimax, mousavi2020black, yang2020off}, SR-DICE requires only a simple convex loss applied to a linear function, after computing the SR. Similar to the deep RL methods which can learn in high-dimensional domains, the SR can be computed easily using behavior-agnostic temporal-difference (TD) methods. This makes our algorithm highly amenable to deep learning architectures and applicable to complex tasks. 

The SR, which measures the expected future occupancy of states for a given policy, has a clear relationship to MIS methods, which estimate the ratio between the occupancy of state-action pairs and the sampling distribution. However, this relationship is muddied in a deep RL context, where the deep SR measures the expected future sum of feature vectors. Our approach, SR-DICE, provides a straightforward and principled method for extracting density ratios from the SR without any modifications to the standard learning procedure of the SR.  
Access to these density ratios is valuable as they have a wide range of possible applications such as policy regularization \citep{nachum2019algaedice, touati2020stable}, imitation learning \citep{kostrikov2019imitation}, off-policy policy gradients~\citep{imani2018off, liu2019off, zhang2019generalized}, non-uniform sampling procedures \citep{sinha2020experience}, or for mitigating distributional shift in offline RL~\citep{fujimoto2018off, kumar2019stabilizing}. 

We highlight the value of the MIS density ratios for one reason in particular--in our theoretical analysis we prove that 
SR-DICE and the deep SR produce \textit{exactly the same value estimate}. This is surprising as SR-DICE takes a distinct approach for value estimation by re-weighting every reward in the dataset with an importance sampling ratio while the deep SR estimates the value in a similar fashion to TD learning. 
This theoretical result extends to the deep RL setting and is consistent in our experimental results. This result is a double-edged sword which (negatively) implies there is no discernible difference of using our MIS approach for policy evaluation, but (positively) implies the estimated density ratios are accurate enough to match the performance of TD methods. This is an important observation as our empirical results demonstrate that previous MIS methods scale very poorly in comparison to TD methods to high dimensions, which is consistent with prior results~\citep{voloshin2019empirical, fu2021benchmarks}. 
Even if there is no difference for OPE, a MIS method which matches the performance of TD-based methods is desirable if we are concerned with estimating the density ratios of the target policy. 

We benchmark the performance of SR-DICE on several high-dimensional domains in MuJoCo \citep{mujoco} and Atari \citep{bellemare2013arcade}, against several recent MIS methods~\citep{nachum2019dualdice, zhang2020gendice}. Our results demonstrate several key findings regarding high-dimensional tasks. 

\textbf{Current MIS methods underperform deep RL at high-dimensional tasks.} While previous results have shown that MIS methods can produce competitive results to TD methods, our empirical results show that MIS methods scale poorly to challenging tasks. In Atari we find that the baseline MIS method exhibit unstable estimates, often reaching errors with many orders of magnitude. Comparatively, the baseline deep RL methods, which rely on TD learning and have a history of achieving high performances in the control setting~\citep{DQN, ppo, fujimoto2018addressing}, outperform the MIS baselines at every task and often by a wide margin.

\textbf{SR-DICE outperforms current MIS methods at policy evaluation and therefore density ratio estimation.} Our empirical results confirm our theoretical analysis, which state that SR-DICE and the standard deep SR approach should produce identical value estimates (with differences due only to changes to the optimization process). While this result may initially sound discouraging, given the direct SR approach is comparable to TD learning, and TD learning significantly outperforms current MIS methods, this result also implies that SR-DICE is a much stronger technique for estimating density ratios than previous methods. 

Ultimately, while SR-DICE produces a similar result to existing deep RL approaches for policy evaluation, it does provide a practical, scalable, and state-of-the-art approach for estimating state-action occupancy density ratios, while highlighting connections between the SR, reward function optimization, and state-action occupancy estimation. For ease of use and reproduction, our code is open-sourced (\url{https://github.com/sfujim/SR-DICE}).

\section{Background}

\textbf{Reinforcement Learning.} RL is a framework for maximizing accumulated reward of an agent interacting with its environment \citep{sutton1998reinforcement}. This problem is typically framed as a Markov Decision Process (MDP) $(\mathcal{S}, \mathcal{A}, \mathcal{R}, p, d_0, \y)$, with state space $\mathcal{S}$, action space $\mathcal{A}$, reward function $\mathcal{R}$, dynamics model $p$, initial state distribution $d_0$ and discount factor $\y$. An agent selects actions according to a policy $\pi:\mathcal{S} \times \mathcal{A} \rightarrow[0,1]$. In this paper we address the problem of off-policy evaluation (OPE) problem where the aim is to measure the normalized expected per-step reward of the policy $R(\pi) = (1 - \y) \E_\pi \lb \sum_{t=0}^\infty \y^t r(s_t,a_t) \rb$. An important notion in OPE is the value function $Q^\pi(s,a) = \E_\pi [ \sum_{t=0}^\infty \y^t r(s_t,a_t) | s_0=s, a_0=a ]$, which measures the expected sum of discounted rewards when following $\pi$, starting from the state-action pair $(s,a)$. 

We define $d^\pi(s,a)$ as the discounted state-action occupancy, the probability of seeing $(s,a)$ under policy $\pi$ with discount $\y$: 
$d^\pi(s,a) = (1 - \y) \sum_{t=0}^\infty \y^t \int_{s_0} d_0(s_0) p_\pi(s_0 \rightarrow s, t) \pi(a|s) ds_0$, 
where $p_\pi(s_0 \rightarrow s, t)$ is the probability of arriving at the state $s$ after $t$ time steps when starting from an initial state $s_0$. This distribution is important as $R(\pi)$ equals the expected reward $r(s,a)$ under $d^\pi$:
\begin{equation}
    R(\pi) = \E_{(s,a) \sim d^\pi, r(s,a)} [ r(s,a) ]. 
\end{equation}

A common approach for estimating $R(\pi)$ is through temporal-difference (TD) learning \citep{sutton1988tdlearning} where an estimate of the value function $Q(s,a)$ is updated over individual transitions $(s,a,r(s,a),s')$ by the following: %
\begin{equation} \label{eqn:tdlearning}
Q(s,a) \leftarrow \al \lp r(s,a) + \y Q(s',a') \rp + (1 - \al) Q(s,a),
\end{equation}
where $a'$ is sampled according to the target policy $\pi$ and $\al$ is the learning rate. Provided an infinite set of transitions, TD learning is known to converge to the true value function in the off-policy setting \citep{jaakkola1994convergence, sutton1998reinforcement}. TD learning can also be applied to other learning problems, such as the successor representation, where the reward $r(s,a)$ in \autoref{eqn:tdlearning} is replaced with the quantity of interest. 

\textbf{Successor Representation.} The successor representation~(SR)~\citep{dayan1993improving} of a policy is a measure of occupancy of future states. It can be viewed as a general value function that learns a vector of the expected discounted visitation for each state. The SR $\Psi^\pi$ of a given policy $\pi$ is defined as $\Psi^\pi(s'|s) = \E_\pi [\sum_{t=0}^\infty \y^t \mathds{1}(s_t=s') | s_0=s ]$. Importantly, the value function can be recovered from the SR by summing over the expected reward of each state $V^\pi(s) = \sum_{s'} \Psi^\pi(s'|s) \E_{a' \sim \pi} [r(s',a')]$. For infinite state and action spaces, the SR can instead be generalized to the expected occupancy over features, known as the deep SR~\citep{kulkarni2016deep} or successor features~\citep{barreto2017successor}. For a given encoding function $\phi: \mathcal{S} \times \mathcal{A} \rightarrow \mathbb{R}^n$, the deep SR $\psi^\pi:\mathcal{S} \times \mathcal{A} \rightarrow \mathbb{R}^n$ is defined as the expected discounted sum of features from the encoding function $\phi$ when following the policy from a given state-action pair: %
\begin{equation}
    \psi^\pi(s,a) = \E_\pi \lb \sum_{t=0}^\infty \y^t \phi(s_t,a_t) \bigg \vert s_0=s, a_0=a \rb.
\end{equation}
If the encoding $\phi(s,a)$ is learned such that the original reward function is a linear function of the encoding $r(s,a)=\mathbf{w}^\top\phi(s,a)$, then similar to the original formulation of SR, the value function can be recovered from a linear function of the SR: $Q^\pi(s,a)=\mathbf{w}^\top \psi^\pi(s,a)$. 
The deep SR network $\psi^\pi$ is trained to minimize the MSE between $\psi^\pi(s,a)$ and $\phi(s,a) + \y \psi'(s',a')$ on transitions $(s,a,s')$ sampled from the dataset. A frozen target network $\psi'$ is used to provide stability~\citep{DQN, kulkarni2016deep}, and is updated to the current network $\psi' \leftarrow \psi^\pi$ after a fixed number of time steps. The encoding function $\phi$ is typically trained by an encoder-decoder network \citep{kulkarni2016deep, machado2017eigenoption, machado2018count}. For OPE where the reward function is learned by minimizing $\lp \mathbf{w}^\top \phi(s,a) - r(s,a) \rp^2$, the SR is comparable to TD learning, as they both estimate the discounted sum of future rewards and use similar updates. 

\textbf{Marginalized Importance Sampling.} Marginalized importance sampling (MIS) is a family of importance sampling approaches for off-policy evaluation in which the performance $R(\pi)$ is evaluated by re-weighting rewards sampled from a dataset $\mathcal{D} = \{(s,a,r,s')\} \sim p(s'|s,a) d^\mathcal{D}(s,a)$, where $d^\mathcal{D}$ is an arbitrary distribution, typically but not necessarily, induced by some behavior policy.  %
It follows that $R(\pi)$ can computed with importance sampling weights on the rewards $\frac{d^\pi(s,a)}{d^{\mathcal{D}}(s,a)}$:
\begin{equation}
R(\pi) = \E_{(s,a) \sim d^\mathcal{D}, r(s,a)} \lb \frac{d^\pi(s,a)}{d^\mathcal{D}(s,a)} r(s,a) \rb.
\end{equation}
The goal of marginalized importance sampling methods is to learn the weights $w(s,a) \approx \frac{d^\pi(s,a)}{d^\mathcal{D}(s,a)}$, using data contained in $\mathcal{D}$.
The main benefit of MIS is that unlike traditional importance methods, the ratios are applied to individual transitions rather than complete trajectories, which can reduce the variance of long or infinite horizon problems. In other cases, the ratios themselves can be used for a variety of applications which require estimating the occupancy of state-action pairs.

\section{A Reward Function Perspective on Distribution Corrections} \label{section:SRDICE}

In this section, we present our behavior-agnostic approach to estimating MIS ratios, called the Successor Representation DIstribution Correction Estimation (SR-DICE). Our main insight is that MIS can be viewed as an optimization over a learned reward function, where the loss is uniquely optimized when the virtual reward is the MIS density ratio. 

Our derived loss function is a straightforward convex loss over the learned reward and the corresponding value function of the target policy. This naturally suggests the use of the successor representation which allows us to maintain an estimate of the value estimate while directly optimizing the reward function. This disentangles the learning process, where the propagation of reward through the MDP can be learned separately from the optimization of the reward. In other words, rather than learn a reward function and value function simultaneously, we tackle each separately, changing the difficult minimax optimization of previous methods into two phases. Interestingly enough, we show that our MIS estimator produces the identical value estimate as traditional deep SR methods. This means the challenging aspect of learning has been pushed onto the computation of the SR, rather than optimizing the density ratio estimate. Fortunately, we can leverage deep RL approaches \citep{DQN, kulkarni2016deep} to make learning the SR stable, giving rise to a practical MIS method for high-dimensional tasks. 

This section begins with the derivation of our core ideas, which shows MIS ratios can be learned through reward function optimization. We then highlight how the SR can be used for reward function optimization in the tabular domain. Finally, we generalize our results to the deep SR setting.

\subsection{Basic Derivation}

In MIS, our aim is to determine the MIS ratios $\frac{d^\pi(s,a)}{d^\mathcal{D}(s,a)}$, using only data sampled from the dataset $\mathcal{D}$ and the target policy $\pi$. This presents a challenge as we have direct access to neither $d^\pi$ nor $d^\mathcal{D}$. 

As a starting point, we begin by following the derivation of DualDICE \citep{nachum2019dualdice}. We first consider the convex function $\frac{1}{2}mx^2 - nx$, which is uniquely minimized by $x^* = \frac{n}{m}$. Now by replacing $x$ with a virtual reward $\hat r(s,a)$, $m$ with the density of the dataset $d^\mathcal{D}(s,a)$, and $n$ with the density of the target policy $d^\pi(s,a)$, we have reformulated the convex function as the following: %
\begin{equation} \label{eqn:ddcore}
\begin{aligned}
    \min_{\hat r(s,a) \forall (s,a)} J(\hat r) :=&~\frac{1}{2} \E_{(s,a) \sim d^\mathcal{D}} \lb 
    \hat r(s,a)^2 \rb \\
    &- \E_{(s,a) \sim d^\pi} \lb \hat r(s,a) \rb.
\end{aligned}
\end{equation}
As \autoref{eqn:ddcore} is still the convex function with renamed variables, following \citet{nachum2019dualdice}, we can observe the following: 

\begin{minipage}{\linewidth}
\begin{observation}
The objective $J(\hat r)$ is minimized when $\hat r(s,a) = \frac{d^\pi(s,a)}{d^\mathcal{D}(s,a)}$ for all state-action pairs $(s,a)$. 
\end{observation}
\end{minipage}

\autoref{eqn:ddcore} is an optimization over two expectations over $d^\mathcal{D}$ and $d^\pi$. While the first expectation over $d^\mathcal{D}$ is tractable by sampling directly from the dataset $\mathcal{D}$, the second expectation relies on the state-action visitation of the target policy $d^\pi(s,a)$ which is not directly accessible without a model of the MDP. At this point, we highlight our choice of notation, $\hat r(s,a)$, in \autoref{eqn:ddcore}. Describing the objective in terms of a fictitious reward $\hat r$ will allow us to draw on familiar relationships between rewards and value functions. %
Consider the equivalence between the value function over initial state-action pairs $(s_0, a_0)$ and the expectation of rewards over the state-action visitation of the policy $(1- \y) \E_{s_0,a_0 \sim \pi}[Q^\pi(s_0,a_0)] = \E_{d^\pi}[r(s,a)]$. It follows that the expectation over $d^\pi$ in \autoref{eqn:ddcore} can be replaced with a value function $\hat Q^\pi$ over~$\hat r$: 
\begin{equation} \label{eqn:ddQ}
\begin{aligned}
    &\min_{\hat r(s,a) \forall (s,a)} J(\hat r) := \frac{1}{2} \E_{(s,a) \sim d^\mathcal{D}} \lb \hat r(s,a)^2 \rb \\
    &\qquad - (1 - \y) \E_{s_0, a_0 \sim \pi(\cdot|s_0)} \lb \hat Q^\pi(s_0,a_0) \rb.
\end{aligned}
\end{equation}
In other words, by noting that the value function is simply the (scaled) expected reward when sampled from the state-action visitation of the target policy, we can replace the impractical expectation over $d^\pi$ with a tractable value function.
This form of the objective, \autoref{eqn:ddQ}, is convenient because we can estimate the expectation over $d^\mathcal{D}$ by sampling directly from the dataset and $\hat Q^\pi$ can be computed using any policy evaluation method. 

While we can estimate both terms in \autoref{eqn:ddQ} with relative ease, the optimization problem is not directly differentiable and would require re-learning the value function $\hat Q^\pi$ with every adjustment to the learned reward $\hat r$. Fortunately, there exists a straightforward paradigm which enables direct reward function optimization known as successor representation (SR). 

\subsection{Tabular SR-DICE}

We will begin by discussing how we can apply the SR to MIS in the tabular setting and then generalize our method to non-linear function approximation afterwards. Consider the relationship between the SR $\Psi^\pi$ of the target policy $\pi$ and its value function: %
\begin{equation} 
\begin{aligned}
&\E_{s_0, a_0 \sim \pi(\cdot|s_0)}[Q^\pi(s_0,a_0)] = \E_{s_0}[V^\pi(s_0)] \\
&\qquad = \E_{s_0} \lb \sum_{s} \Psi^\pi(s|s_0) \E_{a \sim \pi} [r(s,a)] \rb.
\end{aligned}
\end{equation}
It follows that we can create an optimization problem directly over the reward function $\hat r$ by modifying \autoref{eqn:ddQ} to use the SR: 
\begin{equation} \label{eqn:true_sr_dice}
\begin{aligned}
&\min_{\hat r(s,a) \forall (s,a)} J_\Psi(\hat r) := \frac{1}{2} \E_{(s,a) \sim d^\mathcal{D}} \lb \hat r(s,a)^2 \rb \\
&\qquad - (1 - \y) \E_{s_0} \lb \sum_{s} \Psi^\pi(s|s_0) \E_{a \sim \pi} \lb \hat r(s,a) \rb \rb. 
\end{aligned}
\end{equation}

Since this optimization problem is convex, it has a closed form solution. %
The unique optimizer of \autoref{eqn:true_sr_dice} is: 
\begin{equation} \label{eqn:tab_closedform}
\begin{aligned} 
    &(1 - \gamma) \frac{|\mathcal{D}|}{\sum_{(s',a') \in \mathcal{D}} \mathds{1}(s'=s,a'=a)} \\
    &\quad\cdot \E_{s_0} [\pi(a|s) \Psi^\pi(s|s_0)].
\end{aligned}
\end{equation} %

By noting the relationship between the SR and the state occupancy $d^\pi(s,a) = (1 - \y) \E_{s_0} [\Psi^\pi(s|s_0) \pi(s,a)]$ and the fact that $d^\mathcal{D}(s,a) = \frac{\sum_{(s',a') \in \mathcal{D}} \mathds{1}(s'=s,a'=a)}{|\mathcal{D}|}$ we can show this solution simplifies to the MIS density ratio~$\frac{d^\pi(s,a)}{d^\mathcal{D}(s,a)}$.

\begin{theorem}
\autoref{eqn:tab_closedform} is the optimal solution to \autoref{eqn:true_sr_dice} and is equal to $\frac{d^\pi(s,a)}{d^\mathcal{D}(s,a)}$. 
\end{theorem}

A direct consequence of this result is that \autoref{eqn:tab_closedform} can be used with MIS policy evaluation to return the true value estimate $\frac{1}{|\mathcal{D}|} \sum_{(s,a) \in \mathcal{D}} \frac{d^\pi(s,a)}{d^\mathcal{D}(s,a)} r(s,a) = R(\pi)$.

Unfortunately, the form of \autoref{eqn:tab_closedform} relies on the true SR~$\Psi^\pi$, as well as an expectation over $s_0$, both of which may be unobtainable in the setting where we are sampling from a finite dataset $\mathcal{D}$. However, we can still show that with an inexact SR $\hat \Psi$ and sampled estimate of the expectation, using the set of start states $\mathcal{D}_0$ in the dataset, approximating the optimizer \autoref{eqn:tab_closedform} with
\begin{equation}
\begin{aligned}
    r^*(s,a)=&~(1 - \gamma) \frac{|\mathcal{D}|}{\sum_{(s',a') \in \mathcal{D}} \mathds{1}(s'=s,a'=a)} \\
    &\cdot \frac{1}{|\mathcal{D}_0|} \sum_{s_0 \in \mathcal{D}_0} \pi(a|s) \hat \Psi(s|s_0), 
\end{aligned}
\end{equation}
gives an MIS estimator $\frac{1}{|\mathcal{D}|} \sum_{(s,a) \in \mathcal{D}} r^*(s,a) r(s,a)$ of $R(\pi)$ which is identical to the estimate of $R(\pi)$ computed directly with the SR.

\begin{theorem} \label{thm:tab_sr_same}
Let $\bar r(s,a)$ be the average reward in the dataset $\mathcal{D}$ at the state-action pair $(s,a)$. Let $\hat \Psi$ be any approximate SR. The direct SR estimator $(1 - \y) \frac{1}{|\mathcal{D}_0|} \sum_{s_0 \in \mathcal{D}_0} \sum_{s \in \mathcal{S}} \hat \Psi(s|s_0) \sum_{a \in \mathcal{A}} \pi(a|s) \bar r(s,a)$ of $R(\pi)$ is identical to the MIS estimator $\frac{1}{|\mathcal{D}|} \sum_{(s,a) \in \mathcal{D}} r^*(s,a) r(s,a)$. 
\end{theorem}

The take-away is that even when estimating the SR, the approximate density ratio defined by $r^*$ is of sufficiently high quality to match the performance of directly estimating the value with the SR.

\subsection{SR-DICE}

Now we will consider how this MIS estimator can be generalized to continuous states by considering the deep SR $\psi^\pi$ over features $\phi(s,a)$ and optimizing the weights of a linear function $\mathbf{w}$. 

\textbf{SR Refresher.} We begin with a reminder of the details of the deep SR algorithm. The deep SR measures the expected sum of features $\psi^\pi(s,a) = \E_\pi \lb \sum_{t=0}^\infty \y^t \phi(s_t,a_t) \rb$. If the reward can be defined as a linear function over the features $r(s,a) = \mathbf{w}^\top \phi(s,a)$ then the value function can be recovered via a linear function over the deep SR $Q(s,a) = \mathbf{w}^\top \psi^\pi(s,a)$. The typical deep SR pipeline follows three steps: 
\begin{enumerate}[nosep]
\item Learn the encoding $\phi$.
\item Learn the deep SR $\psi^\pi$ over the encoding $\phi$.
\item Learn $\mathbf{w}_\text{SR}$ by minimizing $\lp \mathbf{w}_\text{SR}^\top \phi(s,a) - r(s,a) \rp^2$. 
\end{enumerate}
We leave the first two stages vague as there is flexibility in how they are approached. 
This most commonly involves training the encoding $\phi$ via an encoder-decoder network to reconstruct transitions and training the deep SR $\psi^\pi$ using TD learning-style methods~\citep{kulkarni2016deep, machado2018count}. While we follow this standard practice, specific details are unimportant for our analysis and we relegate implementation-level details to the appendix. %

Given the deep SR $\psi^\pi$, we can use it to learn the MIS ratio. Recall our objective of reward function optimization~(\autoref{eqn:ddQ}). In the deep SR paradigm, both the reward and value function are determined by linear functions with respect to a single weight vector $\mathbf{w}$. 
Consequently, we can modify \autoref{eqn:ddQ} with these linear functions and then optimize the linear weights $\mathbf{w}$ directly: 
\begin{equation} \label{eqn:sr_dice}
\begin{aligned}
    \min_{\mathbf{w}}~&J(\mathbf{w}) := \frac{1}{2} \E_{d^\mathcal{D}} \lb (\mathbf{w}^\top \phi(s,a))^2 \rb \\
    &- (1 - \gamma) \E_{s_0,a_0 \sim \pi(\cdot | s_0)} \lb \mathbf{w}^\top \psi^\pi(s_0,a_0) \rb,
\end{aligned}
\end{equation}
where in practice we replace the expectations with samples from the dataset $\mathcal{D}$ and the subset of start states $\mathcal{D}_0$:
\vspace{-10pt}
\begin{equation} \label{eqn:sr_dice_samples}
\begin{aligned}
    &\min_{\mathbf{w}} J(\mathbf{w}) := \frac{1}{2|\mathcal{D}|} \sum_{(s,a) \in \mathcal{D}} \lb (\mathbf{w}^\top \phi(s,a))^2 \rb \\
    &\quad- (1 - \gamma) \frac{1}{|\mathcal{D}_0|}\sum_{s_0 \in \mathcal{D}_0, a_0} \pi(a_0|s_0) \mathbf{w}^\top \psi^\pi(s_0,a_0).
\end{aligned}
\end{equation}
Again, since the optimization problem \autoref{eqn:sr_dice_samples} is still convex, it has a closed form solution. 
Let $\Phi$ be a $|\mathcal{D}| \times F$ matrix where each row is the feature vector $\phi(s,a)$ with $F$ features. 
Let $\Psi$ be a $|\mathcal{D}_0||\mathcal{A}| \times F$ matrix where each row is the SR weighted by its probability under the policy $\pi(a_0 |s_0) \psi^\pi(s_0,a_0)$. Let $\mathbf{1}$ be a $|\mathcal{D}_0||\mathcal{A}|$ dimensional vector of all $1$. The unique optimizer $\mathbf{w}^*$ of \autoref{eqn:sr_dice_samples} is a $F$ dimensional vector defined as follows: 
\begin{equation} \label{eqn:closedform}
\begin{aligned}
     \mathbf{w}^* = (1 - \y) \frac{|\mathcal{D}|}{|\mathcal{D}_0|} (\Phi^\top \Phi)^{-1} \Psi^\top \mathbf{1}.
\end{aligned}
\end{equation}
In practice, a matrix-based solution is often undesirable and we may prefer iterative, gradient-based solutions for scalability. In this case, we can directly minimize \autoref{eqn:sr_dice_samples} by taking gradient steps with respect to $\mathbf{w}$. 

We now introduce our algorithm Successor Representation stationary DIstribution Correction Estimation (SR-DICE). SR-DICE follows the same first two steps of the standard SR procedure, but replaces the third step with optimizing \autoref{eqn:sr_dice_samples}. Given $\mathbf{w}$, an estimate of $R(\pi)$ can be returned by $\frac{1}{|\mathcal{D}|} \sum_{(s,a,r(s,a)) \in \mathcal{D}} \mathbf{w}^\top \phi(s,a) r(s,a)$, where $\mathbf{w}^\top \phi(s,a) \approx \frac{d^\pi(s,a)}{d^\mathcal{D}(s,a)}$. We summarize SR-DICE in \autoref{algorithm:short}. 

\begin{algorithm}[t]
  \caption{SR-DICE} \label{algorithm:short}
\begin{algorithmic}%
    \STATE \textbf{Input:} SR $\psi$, target network $\psi'$, encoder $\phi$, decoder $D$. 
    \STATE At each time step sample mini-batch of $N$ transitions $(s, a, r, s')$ and start states $s_0$ from $\mathcal{D}$.
	\STATE \textbf{for} $t=1$ \textbf{to} $T_1$ \textbf{do} {\color{blue} \# Encoding $\phi$ loss}%
	\STATE \quad $\min_{\phi, D} \frac{1}{2} (D(\phi(s,a)) - (s,a))^2$. %
	\STATE \textbf{for} $t=1$ \textbf{to} $T_2$ \textbf{do} {\color{blue} \# Deep SR $\psi^\pi$ loss} %
	\STATE \quad $\min_{\psi^\pi} \frac{1}{2} (\phi(s,a) + \y \psi'(s',a') - \psi^\pi(s,a))^2$. %
	\STATE \textbf{for} $t=1$ \textbf{to} $T_3$ \textbf{do} {\color{blue} \# Density ratio $\mathbf{w}$ loss (\autoref{eqn:sr_dice_samples})} %
	\STATE \quad $a_0 \sim \pi(\cdot|s_0)$.
	\STATE \quad $\min_{\mathbf{w}} \frac{1}{2} (\mathbf{w}^\top \phi(s,a))^2 - (1 - \y) \mathbf{w}^\top \psi^\pi(s_0,a_0)$. %
	\STATE \textbf{Output:} $|\mathcal{D}|^{-1} \sum_{(s,a,r) \in \mathcal{D}} \mathbf{w}^\top \phi(s,a) r(s,a) \approx R(\pi)$.
\end{algorithmic}
\end{algorithm}

We now remark upon two important properties of SR-DICE. The first concerns the quality of the quality of the learned MIS ratio. Although it is difficult to make any guarantees on the accuracy of an approximate $\psi^\pi$ trained with deep RL techniques, if we assume $\psi^\pi$ is exact, then we can show that SR-DICE learns the least squares estimator to the desired density ratio. 
\begin{theorem} \label{thm:least_squares} %
If the deep SR is exact, such that $(1 - \y) \E_{s_0,a_0} \lb \psi^\pi(s_0,a_0) \rb = \E_{(s,a) \sim d^\pi} [\phi(s,a)]$, and the support of $d^\pi$ is contained in the dataset $\mathcal{D}$, then the optimizer $\mathbf{w}^*$ of \autoref{eqn:sr_dice_samples}, as defined by \autoref{eqn:closedform}, is the least squares estimator of $\sum_{(s,a) \in \mathcal{D}} \lp \mathbf{w}^\top \phi(s,a) - \frac{d^\pi(s,a)}{d^\mathcal{D}(s,a)} \rp^2$.
\end{theorem}

The take-away from \autoref{thm:least_squares} is that our optimization problem, at least in the idealized setting, produces the same density ratios as directly learning them. This also means that the main source of error in SR-DICE is in the first two phases: learning the encoding $\phi$ and the deep SR $\psi^\pi$. Notably, both of these steps are independent of the main optimization problem of learning $\mathbf{w}$, as we have shifted the challenging aspects of density ratio estimation onto learning the deep SR. This leaves deep RL to do the heavy lifting. The remaining optimization problem, \autoref{eqn:sr_dice}, only involves directly updating the weights of a linear function, and unlike many other MIS methods, requires no tricky minimax optimization. 

The second important property of SR-DICE is that \autoref{thm:tab_sr_same} can be extended to the deep SR setting. That is, when derived from the same approximate SR, the optimal solution to both the SR-DICE estimator and the direct SR estimator produce identical estimates of $R(\pi)$.
\begin{theorem} \label{thm:sr_sr_dice}
Given the least squares estimator $\mathbf{w}_{\text{SR}}$ of $\sum_{(s,a) \in \mathcal{D}} \lp \mathbf{w}^\top \phi(s,a) - r(s,a) \rp^2$ and the optimizer $\mathbf{w}^*$ of \autoref{eqn:sr_dice_samples}, as defined by \autoref{eqn:closedform}, then the traditional SR estimator $\frac{1}{|\mathcal{D}_0|} \sum_{s_0 \in \mathcal{D}_0} \mathbf{w}_\text{SR}^\top \psi^\pi(s_0,a_0)$ of $R(\pi)$ is identical to the SR-DICE estimator $\frac{1}{|\mathcal{D}|} \sum_{(s,a,r(s,a)) \in \mathcal{D}} \mathbf{w}^{*\top} \phi(s,a) r(s,a)$ of $R(\pi)$.
\end{theorem}

This means that SR-DICE produces the same value estimate as the traditional deep SR algorithms, up to errors in the optimization process of $\mathbf{w}$. In other words, SR-DICE does not suffer from the same instability issues that plague other MIS methods when tackling high-dimensional domains where deep RL methods excel (relative to more traditional methods). Although, we typically think of the objective of MIS methods as policy evaluation, since SR-DICE and traditional deep SR produce the same value estimate, there is not a strong argument for using SR-DICE for policy evaluation. However, this also suggests that the estimated density ratios are of reasonably high quality since SR-DICE achieves the same performance as deep RL approaches. Therefore, we can treat SR-DICE is a tractable method for accessing the state-action occupancy of the target policy.

\section{Related Work}

\textbf{Off-Policy Evaluation.} Off-policy evaluation (OPE) is a well-studied problem with several families of approaches. 
One family of approaches is based on importance sampling, which re-weights trajectories by the ratio of likelihoods under the target and behavior policy \citep{precup2001off}. Importance sampling methods are unbiased but suffer from variance which can grow exponentially with the length of trajectories \citep{li2015toward, jiang2016doubly}. Consequently, research has focused on variance reduction \citep{thomas2016data, munos2016safe, farajtabar2018more} or contextual bandits~\citep{dudik2011doubly, wang2017optimal}. Marginalized importance sampling methods~\citep{liu2018breaking} aim to avoid this exponential variance by considering the ratio in stationary distributions, giving an estimator with variance which is polynomial with respect to horizon~\citep{xie2019towards, liu2019understanding}. Follow-up work has introduced a variety of approaches and improvements, allowing them to be behavior-agnostic~\citep{nachum2019dualdice, uehara2019minimax, mousavi2020black, yang2020off} and operate in the undiscounted setting~\citep{zhang2020gendice, zhang2020gradientdice}. In a similar vein, some OPE methods rely on emphasizing, or re-weighting, updates based on their stationary distribution~\citep{sutton2016emphatic, mahmood2017multi, hallak2017consistent, gelada2019off}, or learning the stationary distribution directly~\citep{wang2007dual, wang2008stable}.

For many deep RL algorithms \citep{DQN, DDPG}, off-policy evaluation is based on TD learning \citep{sutton1988tdlearning} and approximate dynamic programming techniques such as Fitted Q-Iteration \citep{ernst2005tree, riedmiller2005neural, yang2019theoretical}. While empirically successful, these approaches lose any theoretical guarantees with non-linear function approximation \citep{tsitsiklis1997analysis, chen2019information}. Regardless, they have been shown to achieve a high performance at benchmark OPE tasks~\citep{voloshin2019empirical, fu2021benchmarks}.

\textbf{Successor Representation.} Introduced originally by \citet{dayan1993improving} as an approach for improving generalization in temporal-difference methods, successor representations~(SR) were revived by recent work on deep successor RL~\citep{kulkarni2016deep} and successor features \citep{barreto2017successor} which demonstrated that the SR could be generalized to a function approximation setting. The SR has found applications for task transfer~\citep{barreto2018transfer, grimm2019disentangled}, navigation \citep{zhang2017deep, zhu2017visual}, and exploration \citep{machado2018count, janz2019successor}. It has also been used in a neuroscience context to model generalization and human reinforcement learning~\citep{gershman2012successor, momennejad2017successor, gershman2018successor}. The SR and our work also relate to state representation learning~\citep{lesort2018state} and general value functions~\citep{sutton2005temporal, sutton2011horde}.

\section{Experiments}

\begin{figure*}[t]
    \centering
    \includegraphics[width=\linewidth]{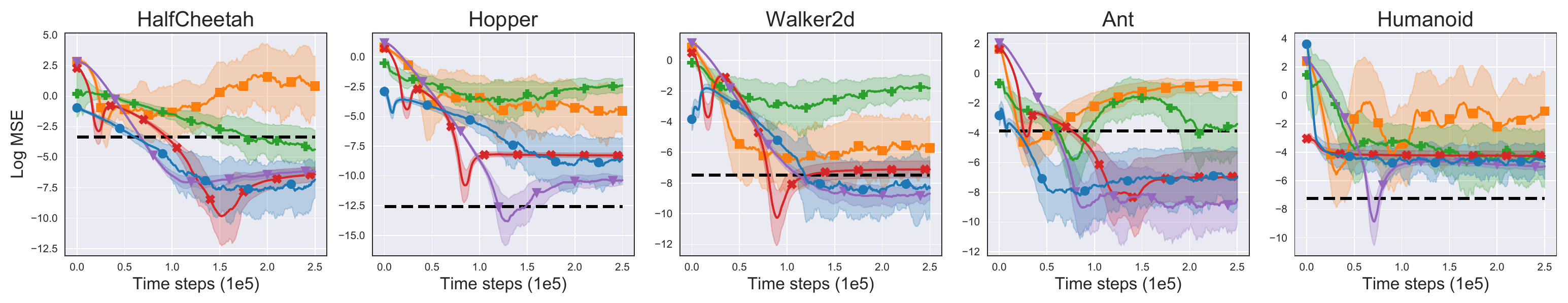}
    \small{
    \ccirc{31}{119}{180} SR-DICE \quad
    \csquare{255}{127}{14} DualDICE \quad
    \cplus{44}{160}{44} GradientDICE \quad
    \ctimes{214}{39}{40} Deep SR \quad 
    \ctriangle{148}{103}{189} Deep TD \quad
    \cdashed Behavior $R(\pi_b)$
    }
    \caption{Off-policy evaluation results on the continuous action MuJoCo domain using the \textit{easy} experimental setting (500k time steps and $\sigma_b=0.133$), matching the setting of previous methods~\citep{zhang2020gendice}. The shaded area captures one standard deviation across 10 trials. We remark that this setting can be considered easy as the behavior policy achieves a lower error, often outperforming all agents. SR-DICE significantly outperforms the other MIS methods on all environments, except for Humanoid, where GradientDICE achieves a comparable performance.}
    \label{fig:mujoco_easy}
\end{figure*}

\begin{figure*}[t]
    \centering
    \includegraphics[width=\linewidth]{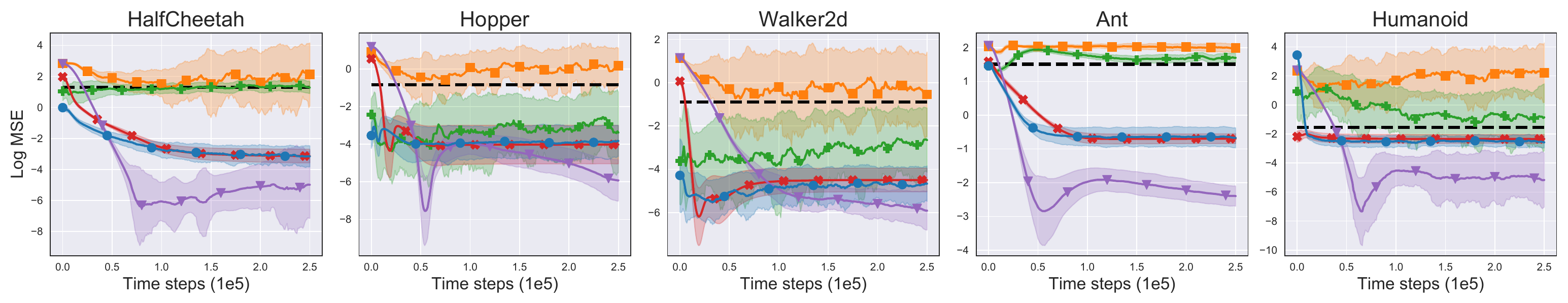}
    \small{
    \ccirc{31}{119}{180} SR-DICE \quad
    \csquare{255}{127}{14} DualDICE \quad
    \cplus{44}{160}{44} GradientDICE \quad
    \ctimes{214}{39}{40} Deep SR \quad 
    \ctriangle{148}{103}{189} Deep TD \quad
    \cdashed Behavior $R(\pi_b)$
    }
    \caption{Off-policy evaluation results on the continuous action MuJoCo domain using the \textit{hard} experimental setting (50k time steps, $\sigma_b=0.2$, random actions with $p=0.2$). The shaded area captures one standard deviation across 10 trials. This setting uses significantly fewer time steps than the \textit{easy} setting and the behavior policy is a poor estimate of the target policy. Again, we see SR-DICE outperforms the MIS methods, demonstrating the benefits of our proposed decomposition and simpler optimization. This setting also shows the benefits of deep RL methods over MIS methods for OPE in high-dimensional domains, as deep TD performs the strongest in every environment.}
    \label{fig:mujoco_results}
\end{figure*}

\begin{figure*}[t]
    \centering
    \includegraphics[width=\linewidth]{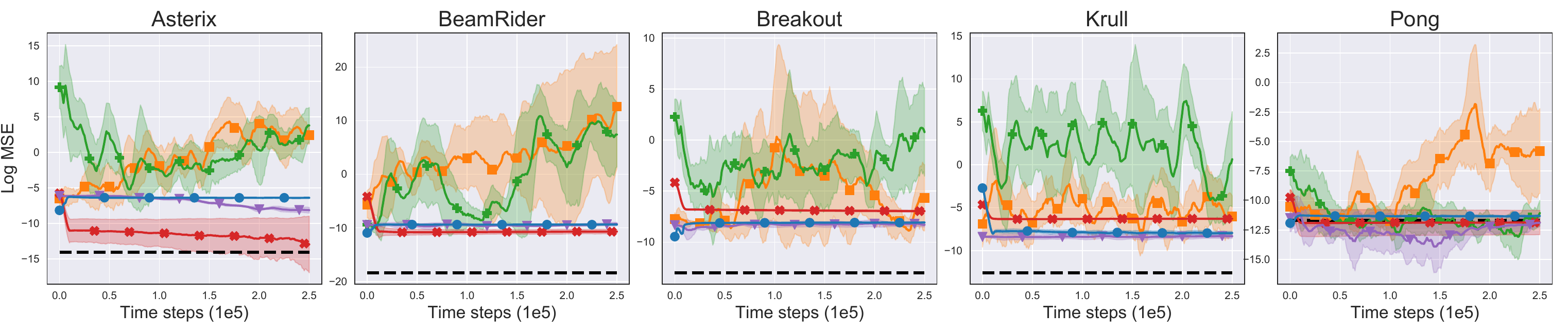}
    \small{
    \ccirc{31}{119}{180} SR-DICE \quad
    \csquare{255}{127}{14} DualDICE \quad
    \cplus{44}{160}{44} GradientDICE \quad
    \ctimes{214}{39}{40} Deep SR \quad 
    \ctriangle{148}{103}{189} Deep TD \quad
    \cdashed Behavior $R(\pi_b)$
    }
    \caption{The log MSE for off-policy evaluation in the image-based Atari domain. This high-dimensional domain tests the ability of each method to scale to more complex environments. The shaded area captures one standard deviation across 3 trials. We can see the MIS baselines diverge on this challenging environment, while the remaining methods perform similarly. Perhaps surprisingly, on most games, the na\"ive baseline of using $R(\pi_b)$ from the behavior policy outperforms all methods by a fairly significant margin. Although the estimates from deep RL methods are stable, they are biased, resulting in a higher MSE.}
    \label{fig:atari_results}
\end{figure*}

\begin{figure*}
\begin{minipage}[b]{0.45\textwidth}
\includegraphics[width=\textwidth]{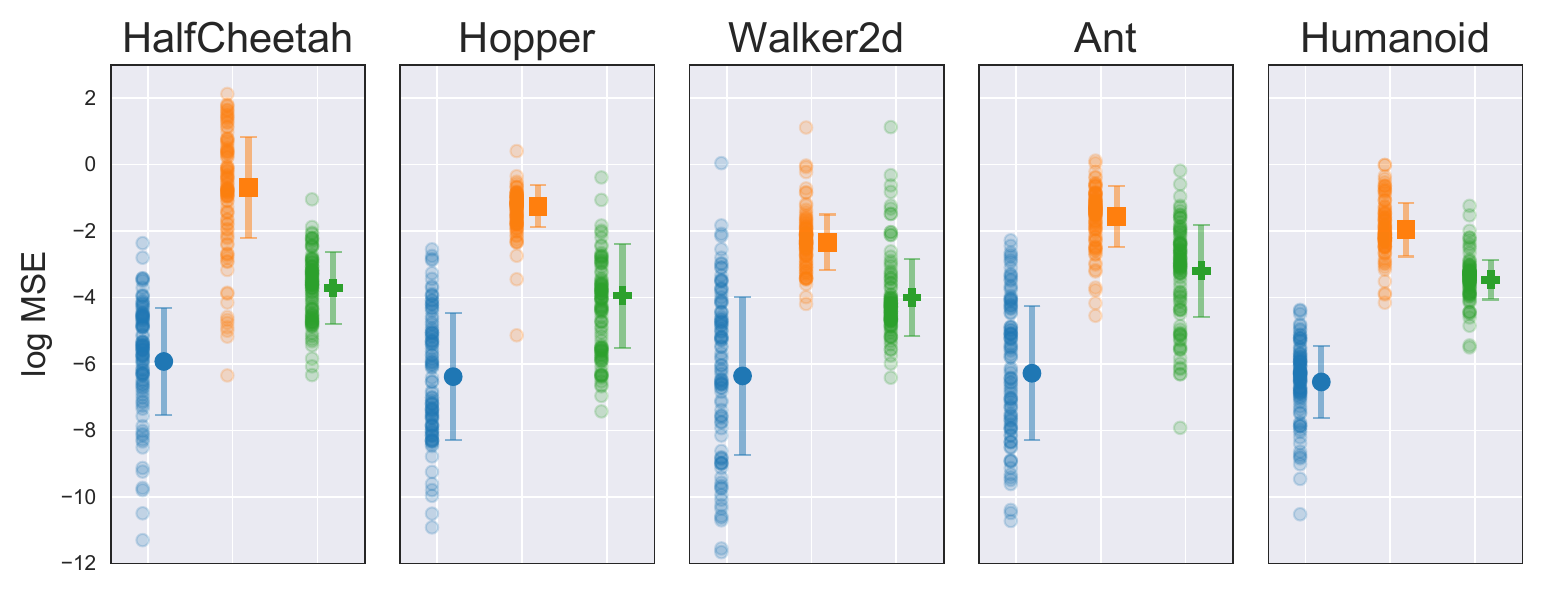}

\quad\small{
\ccirc{31}{119}{180} SR-DICE 
\csquare{255}{127}{14} DualDICE
\cplus{44}{160}{44} GradientDICE
}
\end{minipage}
\hfill
\begin{minipage}[b]{0.55\textwidth}
\small
\begin{tabular}{lrrr}
\toprule
            & SR-DICE & DualDICE & GradientDICE \\
\midrule
HalfCheetah & \textbf{-5.9$\pm$1.6 (82.6)}   & -0.7$\pm$1.5 (0.4)     & -3.7$\pm$1.1 (17.1)        \\
Hopper      & \textbf{-6.4$\pm$1.9 (80.4)}   & -1.3$\pm$0.6 (0.5)     & -3.9$\pm$1.6 (19.1)        \\
Walker2d    & \textbf{-6.4$\pm$2.4 (81.7)}   & -2.3$\pm$0.8 (0.5)     & -4.0$\pm$1.2 (17.8)         \\
Ant         & \textbf{-6.3$\pm$2.0 (81.9)}  & -1.6$\pm$0.9 (1.4)     & -3.2$\pm$1.4 (16.7)        \\
Humanoid    & \textbf{-6.5$\pm$1.1 (98.8)}   & -2.0$\pm$0.8 (0.0)     & -3.5$\pm$0.6 (1.2)         \\
\bottomrule
\end{tabular}
\vspace{15pt}
\end{minipage}

\vspace{-10pt}
\subfloat[Error Visualization]{\hspace{0.45\linewidth}}\hfill
\subfloat[Log MSE \& (Percentage of rewards functions with minimum error)]{\hspace{0.55\linewidth}}
\caption{To evaluate the quality of the MIS ratios, we evaluate each MIS ratio with $1000$ randomly sampled reward functions and compare to the ground truth on-policy value estimates. (Left) Visualization of the distribution of error. Only $100$ points are displayed for visual clarity. Error bars are over the standard deviation. To normalize values across rewards functions, we divide both the estimate and ground truth of $R(\pi)$ by the average reward in the dataset. 
(Right) Average log MSE and the standard deviation. In brackets is the percentage of reward functions where each method achieves the lowest error. We can see that SR-DICE achieves a low log MSE over a wide range of reward functions and outperforms the competing MIS methods on a high percentage of reward functions.} \label{fig:random}
\end{figure*}

\begin{figure*}[t]
\centering
\includegraphics[width=0.225\linewidth]{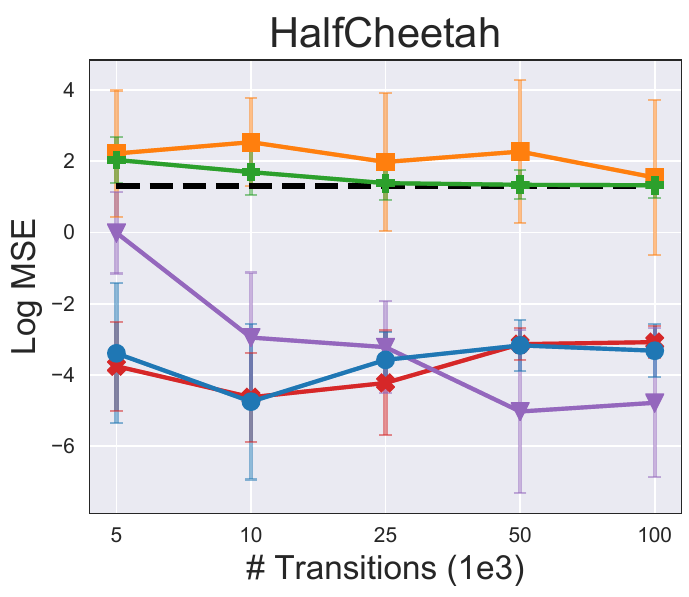}\hfill
\includegraphics[width=0.225\linewidth]{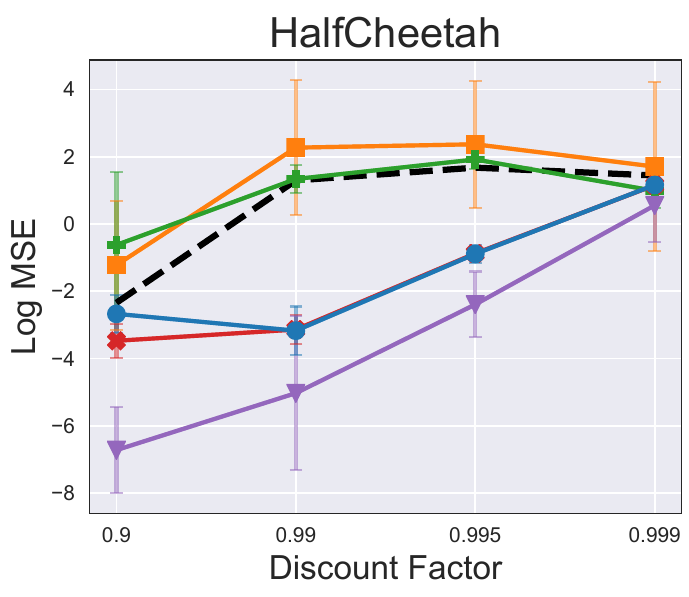}\hfill
\includegraphics[width=0.225\linewidth]{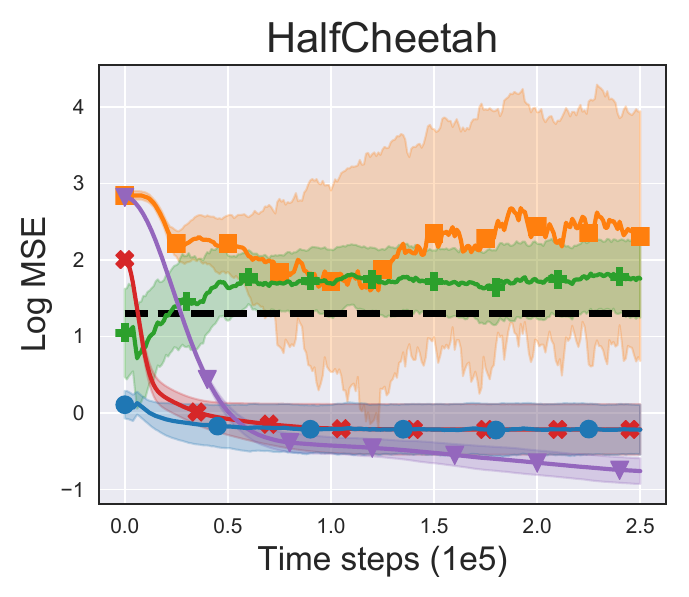}\hfill
\includegraphics[width=0.225\linewidth]{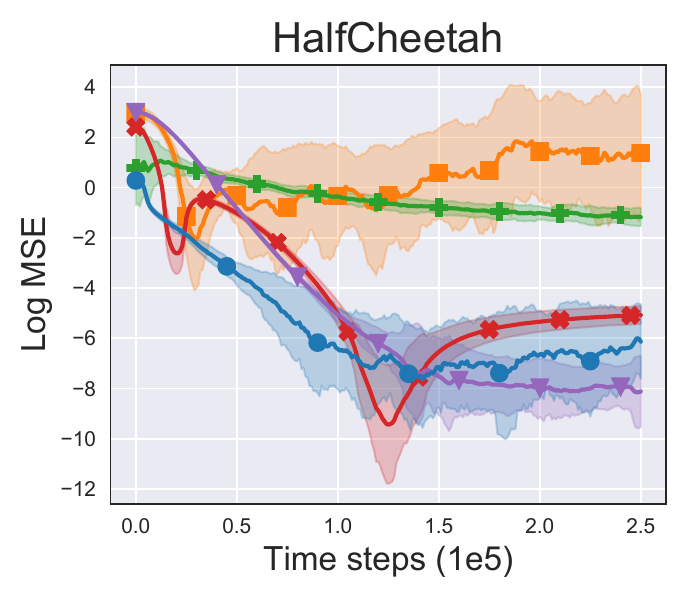}

\small{
\ccirc{31}{119}{180} SR-DICE \quad
\csquare{255}{127}{14} DualDICE \quad
\cplus{44}{160}{44} GradientDICE \quad
\ctimes{214}{39}{40} Deep SR \quad 
\ctriangle{148}{103}{189} Deep TD \quad
\cdashed Behavior $R(\pi_b)$
}
\vspace{-2mm}

\subfloat[dataset size]{\hspace{0.225\linewidth}}\hfill
\subfloat[Discount factor $\y$]{\hspace{0.225\linewidth}}\hfill
\subfloat[Increased noise]{\hspace{0.225\linewidth}}\hfill
\subfloat[Deterministic policies]{\hspace{0.23\linewidth}}

\caption{Ablation study results for the HalfCheetah task. We default to the \textit{hard} setting wherever possible. Error bars and the shaded area captures one standard deviation over 10 trials. (a) We vary the size of the dataset $\mathcal{D}$. (b) We vary the discount factor $\y$. (c) We use a new behavior policy with $\N(0,\sigma_b^2)$ noise with $\sigma_b=0.5$. (d) We use the same deterministic behavior and target policy.}
\label{fig:ablation}
\end{figure*}

To evaluate our method, we perform several off-policy evaluation (OPE) experiments on a variety of domains. The aim is to evaluate the normalized average discounted reward $\E_{(s,a) \sim d^\pi, r} [r(s,a)]$ of a target policy $\pi$. We benchmark our algorithm against two MIS methods, DualDICE~\citep{nachum2019dualdice} and GradientDICE~\citep{zhang2020gradientdice}, two deep RL approaches and the true return of the behavior policy. The first deep RL method is a DQN-style approach~\citep{DQN} where actions are selected by $\pi$ (denoted Deep TD) and the second is the deep SR where the weight $\mathbf{w}$ is trained to minimize the MSE between $\mathbf{w}^\top \phi(s,a)$ and~$r(s,a)$~\citep{kulkarni2016deep}. 
Environment-specific experimental details are presented below, and complete algorithmic and hyper-parameter details are included in the appendix. %

\textbf{Continuous Action Experiments.} We evaluate the methods on a variety of MuJoCo environments \citep{OpenAIGym, mujoco}. %
We examine two experimental settings. In both settings the target policy $\pi$ and behavior policy $\pi_b$ are stochastic versions of a deterministic policy $\pi_d$ obtained from training the TD3 algorithm \citep{fujimoto2018addressing}. We evaluate a target policy $\pi = \pi_d + \N(0,\sigma^2)$, where $\sigma=0.1$. 
\begin{itemize}[nosep, leftmargin=*]
    \item For the \textit{easy} setting, we gather a dataset of 500k transitions using a behavior policy $\pi_b = \pi_d + \N(0,\sigma_b^2)$, where $\sigma_b=0.133$. This setting roughly matches the experimental setting used by GradientDICE~\citet{zhang2020gendice}. 
    \item For the \textit{hard} setting, we gather a significantly smaller dataset of 50k transitions using a behavior policy which acts randomly with $p=0.2$ and uses $\pi_d + \N(0,\sigma_b^2)$, where $\sigma_b=0.2$, with $p=0.8$. 
\end{itemize}
Unless specified otherwise, we use a discount factor of $\y=0.99$ and all hyper-parameters are kept constant across environments. All experiments are performed over $10$ seeds. We display the results of the \textit{easy} setting in \autoref{fig:mujoco_easy} and the \textit{hard} setting in \autoref{fig:mujoco_results}.

\textbf{Atari Experiments.} 
To demonstrate our approach can scale to even more complex domains, we perform experiments with several Atari games \citep{bellemare2013arcade}, which are challenging due to their high-dimensional image-based state space. 
Standard pre-processing steps are applied~\citep{castro2018dopamine} and sticky actions are used \citep{machado2018revisiting} to increase difficulty and remove determinism. Each method is trained on a dataset of one million time steps. The target policy is the deterministic greedy policy trained by Double DQN \citep{DoubleDQN}. The behavior policy is the $\e$-greedy policy with $\e=0.1$. We use a discount factor of $\y = 0.99$. Experiments are performed over 3 seeds. Results are displayed in \autoref{fig:atari_results}. Additional experiments with different behavior policies can be found in the appendix. 

\textbf{Evaluating the MIS ratios.} To evaluate the quality of the MIS ratios themselves, we perform a randomized reward experiment. As the MIS ratio is only the value $w$ that will return the true value of $R(\pi) = \E_\mathcal{D} [ w \cdot r(s,a) ]$ for all possible reward functions~\citep{uehara2019minimax}, we generate a large set of rewards functions with a randomly-initialized neural network, and evaluate the estimate of $R(\pi)$ obtained from each MIS method on each reward function. The ground-truth is estimated by a set of $100$ on-policy trajectories generated by $\pi$. We generate $1000$ reward functions, with scalar values in the range $[0,10]$ and remove any redundant reward functions from the set. The MIS ratios and dataset are taken from the \textit{hard} setting. Experiments are performed over $5$ seeds. We report the results in \autoref{fig:random}. 

\textbf{Discussion.} Across the board we find SR-DICE significantly outperforms the MIS methods. 
Looking at the estimated values of $R(\pi)$ in the continuous action environments, \autoref{fig:mujoco_results}, we can see that SR-DICE converges rapidly and maintains a stable estimate, while the MIS methods are particularly unstable, especially in the case of DualDICE. These observations are consistent in the Atari domain (\autoref{fig:atari_results}). In accordance with our theoretical analysis, Deep SR and SR-DICE perform similarly in every task, further suggesting that the limiting factor in SR-DICE is the quality of the deep successor representation, rather than learning the density ratios. In the randomized reward experiment, we find that SR-DICE vastly outperforms the other MIS methods in average log MSE, and compares favorably against the other MIS methods in over $80$\% of reward functions. In the most challenging task, Humanoid, SR-DICE is the best method in over $98$\% of reward functions. This suggests that SR-DICE provides much higher quality MIS ratio estimates than previous methods.

\textbf{Ablation.} To study the robustness of SR-DICE relative to the competing methods, we perform an ablation study and investigate the effects of dataset size, discount factor, and two different behavior policies. Unless specified otherwise, we use experimental settings matching the \textit{hard} setting. We report the results in \autoref{fig:ablation}. In the dataset size experiment (a), SR-DICE perform well with as few as $5$k transitions ($5$ trajectories). In some instances, the performance is unexpectedly improved with less data, although incrementally. For small datasets, the SR methods outperform Deep TD. One hypothesis is that the encoding acts as an auxiliary reward and helps stabilize learning in the low data regime. In (b) we report the performance over changes in discount factor. The relative ordering across methods is unchanged. 
In (c) we use a behavior policy of $\N(0, \sigma_b^2)$, with $\sigma_b=0.5$, a much larger standard deviation than either setting for continuous control. 
The results are similar to the original setting, with an increased bias on the deep RL methods. In (d) we use the underlying deterministic policy as both the behavior and target policy. Even though this setup should be easier since the task is no longer off-policy, the baseline MIS methods perform surprisingly poorly, once again demonstrating their weakness on high-dimensional domains.

\section{Conclusion}

In this paper, we introduce a method which can perform marginalized importance sampling (MIS) using the successor representation (SR) of the target policy. This is achieved by deriving an MIS formulation that can be viewed as reward function optimization. 
By using the SR, we effectively disentangle the dynamics of the environment from learning the reward function. This allows us to (a) use well-known deep RL methods to effectively learn the SR in challenging domains \citep{DQN, kulkarni2016deep} and (b) provide a straightforward loss function to learn the density ratios without any optimization tricks necessary for previous methods \citep{liu2018breaking, uehara2019minimax, nachum2019dualdice, zhang2020gradientdice, yang2020off}. 
Our resulting algorithm, SR-DICE, outperforms prior MIS methods in terms of both performance and stability and is the first MIS method which demonstrably scales to high-dimensional problems.

\section{Acknowledgements}
Scott Fujimoto is supported by a NSERC scholarship as well as the Borealis AI Global Fellowship Award. This research was enabled in part by support provided by Calcul Qu\'ebec and Compute Canada. We would like to thank Wesley Chung, Pierre-Luc Bacon, Edward Smith, and Wei-Di Chang for helpful discussions and feedback.

\bibliography{example_paper}
\bibliographystyle{icml2021}

\clearpage

\setcounter{theorem}{0}
\setcounter{corollary}{0}
\setcounter{observation}{0}
\setcounter{lemma}{0}

\onecolumn

\appendix

\section{Detailed Proofs.} \label{appendix:section:proofs}

\subsection{Observation 1}

\begin{observation} \label{appendix:theorem4}
The objective $J(\hat r)$ is minimized when $\hat r(s,a) = \frac{d^\pi(s,a)}{d^\mathcal{D}(s,a)}$ for all state-action pairs $(s,a)$. 
\end{observation}

Where
\begin{align}
    \min_{\hat r(s,a) \forall (s,a)} J(\hat r)~&:= \frac{1}{2} \E_{(s,a) \sim d^\mathcal{D}} \lb \hat r(s,a)^2 \rb - (1 - \y) \E_{s_0,a_0} \lb \hat Q^\pi(s_0,a_0) \rb \\
    &= \frac{1}{2} \E_{(s,a) \sim d^\mathcal{D}} \lb \hat r(s,a)^2 \rb - \E_{(s,a) \sim d^\pi} \lb \hat r(s,a) \rb.
\end{align}

\textit{Proof.} 

Take the partial derivative of $J(\hat r)$ with respect to $\hat r(s,a)$:
\begin{equation}
\frac{\partial}{\partial \hat r(s,a)} \lp \frac{1}{2} \E_{(s,a) \sim d^\mathcal{D}} \lb \hat r(s,a)^2 \rb - \E_{(s,a) \sim d^\pi} \lb \hat r(s,a) \rb \rp = d^\mathcal{D}(s,a) \hat r(s,a) - d^\pi(s,a).
\end{equation}
Then setting $\frac{\partial J(\hat r)}{\partial \hat r(s,a)} = 0$, we have that $J(\hat r)$ is minimized when $\hat r(s,a) = \frac{d^\pi(s,a)}{d^\mathcal{D}(s,a)}$ for all state-action pairs $(s,a)$.

\hfill $\blacksquare$

\subsection{Theorem 1}

\begin{theorem} \label{appendix:thm1}
\autoref{appendix:eqn:tab_closedform} is the optimal solution to \autoref{appendix:eqn:true_sr_dice} and is equal to $\frac{d^\pi(s,a)}{d^\mathcal{D}(s,a)}$. 
\end{theorem}

Where
\begin{equation} \label{appendix:eqn:true_sr_dice}
\min_{\hat r(s,a) \forall (s,a)} J_\Psi(\hat r) := \frac{1}{2} \E_{(s,a) \sim d^\mathcal{D}} \lb \hat r(s,a)^2 \rb - (1 - \y) \E_{s_0} \lb \sum_{s} \Psi^\pi(s|s_0) \E_{a \sim \pi} \lb \hat r(s,a) \rb \rb.
\end{equation}
and 
\begin{equation} \label{appendix:eqn:tab_closedform}
    \hat r^* (s,a)= (1 - \gamma) \frac{|\mathcal{D}|}{\sum_{(s',a') \in \mathcal{D}} \mathds{1}(s'=s,a'=a)} \E_{s_0} [\pi(a|s) \Psi^\pi(s|s_0)]. 
\end{equation}

\textit{Proof.}

First we consider the relationship between \autoref{appendix:eqn:true_sr_dice} and \autoref{appendix:eqn:tab_closedform}. Take the gradient of \autoref{appendix:eqn:true_sr_dice}:
\begin{equation}
\nabla_{\hat r(s,a)} J_\Psi(\hat r) := d^\mathcal{D}(s,a) \hat r(s,a) - (1 - \gamma) \E_{s_0} [ \Psi^\pi(s|s_0) \pi(a|s) ].
\end{equation}

Where we can replace $d^\mathcal{D}(s,a)$ with $\frac{1}{|\mathcal{D}|} \sum_{(s',a') \in \mathcal{D}} \mathds{1}(s'=s,a'=a)$:
\begin{equation}
\nabla_{\hat r(s,a)} J_\Psi(\hat r) := \frac{1}{|\mathcal{D}|} \lp \sum_{(s',a') \in \mathcal{D}} \mathds{1}(s'=s,a'=a) \rp \hat r(s,a) - (1 - \gamma) \E_{s_0} [ \Psi^\pi(s|s_0) \pi(a|s) ].
\end{equation}

Setting the above gradient equal to 0 and solving for $\hat r(s,a)$ we have the optimizer of $J_\Psi(\hat r)$.
\begin{equation}
    \hat r^*(s,a) = (1 - \gamma) \frac{|\mathcal{D}|}{\sum_{(s',a') \in \mathcal{D}} \mathds{1}(s'=s,a'=a)} \E_{s_0} [ \pi(a|s) \Psi^\pi(s|s_0) ]. 
\end{equation}

Now consider the relationship between \autoref{appendix:eqn:tab_closedform} and $\frac{d^\pi(s,a)}{d^\mathcal{D}(s,a)}$. Recall the definition of the state occupancy $d^\pi(s,a)$:
\begin{align}
    d^\pi(s,a)~&= (1 - \y) \sum_{t=0}^\infty \y^t \int_{s_0} d_0(s_0) p_\pi(s_0 \rightarrow s, t) \pi(a|s) ds_0 \\
    &= (1 - \y) \E_{s_0} \lb \sum_{t=0}^\infty \y^t p_\pi(s_0 \rightarrow s, t) \pi(a|s) \rb.
\end{align}

Now consider the SR:
\begin{align}
\Psi^\pi(s|s_0)~&= \E_\pi \lb \sum_{t=0}^\infty \y^t \mathds{1}(s_t=s) | s_0 \rb \\
&= \sum_{t=0}^\infty \y^t p_\pi(s_0 \rightarrow s, t).
\end{align}

It follows that the SR and the state occupancy share the relationship: 
\begin{equation}
d^\pi(s,a) = (1 - \y) \E_{s_0} [\Psi^\pi(s|s_0) \pi(s,a)].    
\end{equation}

Finally recall that $d^\mathcal{D}(s,a) = \frac{1}{|\mathcal{D}|} \sum_{(s',a') \in \mathcal{D}} \mathds{1}(s'=s,a'=a)$. Note in this case, the relationship is exact and does not rely on an expectation. It follows that:
\begin{align}
    \hat r^*(s,a) &= (1 - \gamma) \frac{|\mathcal{D}|}{\sum_{(s',a') \in \mathcal{D}} \mathds{1}(s'=s,a'=a)} \E_{s_0} [ \pi(a|s) \Psi^\pi(s|s_0) ] \\
    &= \frac{d^\pi(s,a)}{d^\mathcal{D}(s,a)}.
\end{align}

\hfill $\blacksquare$

\subsection{Theorem 2}

\begin{theorem}
Let $\bar r(s,a)$ be the average reward in the dataset $\mathcal{D}$ at the state-action pair $(s,a)$. Let $\hat \Psi$ be any approximate SR. The direct SR estimator $(1 - \y) \frac{1}{|\mathcal{D}_0|} \sum_{s_0 \in \mathcal{D}_0} \sum_{s \in \mathcal{S}} \hat \Psi(s|s_0) \sum_{a \in \mathcal{A}} \pi(a|s) \bar r(s,a)$ of $R(\pi)$ is identical to the MIS estimator $\frac{1}{|\mathcal{D}|} \sum_{(s,a) \in \mathcal{D}} r^*(s,a) r(s,a)$. 
\end{theorem}

\textit{Proof.}

First we will derive the SR estimator for $R(\pi)$. Define $\bar r(s,a)$ as the average of all $r(s,a)$ in the dataset $\mathcal{D}$:
\begin{equation}
    \bar r(s,a) = \begin{cases}
        \sum_{r(s,a) \in \mathcal{D}} \frac{r(s,a)}{\sum_{(s',a') \in \mathcal{D}} \mathds{1}(s'=s,a'=a)} & \text{if } (s,a) \in \mathcal{D} \\
        0 & \text{otherwise.} 
    \end{cases}
\end{equation}

Recall by definition $V^\pi(s) = \sum_{s'} \Psi^\pi(s'|s) \E_{a' \sim \pi} [r(s',a')]$. It follows that  $\sum_{s'} \Psi^\pi(s'|s) \sum_{a' \in \mathcal{A}} \pi(a'|s') \bar r(s',a')$ is an unbiased estimator of $V^\pi(s)$. It follows that estimating $R(\pi) = (1 - \y) \E_{s_0} [V^\pi(s_0)]$ with the SR would be computed by 
\begin{equation}
(1 - \y) \frac{1}{|\mathcal{D}_0|} \sum_{s_0 \in \mathcal{D}_0} \sum_{s \in \mathcal{S}} \Psi^\pi(s|s_0) \sum_{a \in \mathcal{A}} \pi(a|s) \bar r(s,a).
\end{equation}

Now consider the SR-DICE estimator for $R(\pi)$. By expanding and simplifying we arrive at the SR estimator for $R(\pi)$:
\begin{align}
    &\frac{1}{|\mathcal{D}|} \sum_{(s,a,r(s,a)) \in \mathcal{D}} r^*(s,a) r(s,a) \\
    &= \frac{1}{|\mathcal{D}|} \sum_{(s,a,r(s,a)) \in \mathcal{D}} (1 - \y) \frac{|\mathcal{D}|}{\sum_{(s',a') \in \mathcal{D}} \mathds{1}(s'=s,a'=a)} \frac{1}{|\mathcal{D}_0|} \sum_{s_0 \in \mathcal{D}_0} \pi(a|s) \Psi^\pi(s|s_0) r(s,a) \\
    &= (1 - \y) \frac{1}{|\mathcal{D}_0|} \sum_{s_0 \in \mathcal{D}_0} \sum_{(s,a,r(s,a)) \in \mathcal{D}} \Psi^\pi(s|s_0) \pi(a|s) \frac{1}{\sum_{(s',a') \in \mathcal{D}} \mathds{1}(s'=s,a'=a)}  r(s,a) \\
    &= (1 - \y) \frac{1}{|\mathcal{D}_0|} \sum_{s_0 \in \mathcal{D}_0} \sum_{s \in \mathcal{S}} \Psi^\pi(s|s_0) \sum_{a \in \mathcal{A}} \pi(a|s) \bar r(s,a).
\end{align}

\hfill $\blacksquare$

\subsection{Derivation of $\mathbf{w}^*$} \label{appendix:sec:wstar}

Recall the optimization objective $J(\mathbf{w})$:
\begin{equation} \label{appendix:eqn:jw3}
\min_{\mathbf{w}} J(\mathbf{w}) := \frac{1}{2|\mathcal{D}|} \sum_{(s,a) \in \mathcal{D}} \lb (\mathbf{w}^\top \phi(s,a))^2 \rb - (1 - \gamma) \frac{1}{|\mathcal{D}_0|}\sum_{s_0 \in \mathcal{D}_0, a_0} \pi(a_0|s_0) \mathbf{w}^\top \psi^\pi(s_0,a_0).
\end{equation}

Let: 
\begin{itemize}[nosep]
    \item $\Phi$ be a $|\mathcal{D}| \times F$ matrix where each row is the feature vector $\phi(s,a)$ with $F$ features.
    \item  $\Psi$ be a $|\mathcal{D}_0||\mathcal{A}| \times F$ matrix where each row is $\pi(a_0 |s_0) \psi^\pi(s_0,a_0)$, the SR weighted by its probability under the policy over all states $s_0$ in dataset of start states $\mathcal{D}_0$ and all actions. 
    \item $\mathbf{1}$ be a $|\mathcal{D}_0||\mathcal{A}|$ dimensional vector of all $1$.
\end{itemize}

We can reformulate \autoref{appendix:eqn:jw3} as the following:
\begin{equation}
    \frac{1}{2 |\mathcal{D}|}(\Phi \mathbf{w})^\top \Phi \mathbf{w} - (1 - \y) \frac{1}{|\mathcal{D}_0|} \mathbf{1}^\top \Psi \mathbf{w}.
\end{equation}

Now take the gradient:
\begin{align}
    &\g_\mathbf{w} \lp \frac{1}{2 |\mathcal{D}|}(\Phi \mathbf{w})^\top \Phi \mathbf{w} - (1 - \y) \frac{1}{|\mathcal{D}_0|} \mathbf{1}^\top \Psi \mathbf{w} \rp \\
    &= \frac{1}{|\mathcal{D}|} \mathbf{w}^\top \Phi^\top \Phi - (1 - \y) \frac{1}{|\mathcal{D}_0|} \mathbf{1}^\top \Psi
\end{align}

And set it equal to 0 to solve for $\mathbf{w}^*$:
\begin{align}
    \frac{1}{|\mathcal{D}|} \mathbf{w}^\top \Phi^\top \Phi~&- (1 - \y) \frac{1}{|\mathcal{D}_0|} \mathbf{1}^\top \Psi = 0 \\
    \frac{1}{|\mathcal{D}|} \mathbf{w}^\top \Phi^\top \Phi~&= (1 - \y) \frac{1}{|\mathcal{D}_0|} \mathbf{1}^\top \Psi \\
    \Phi^\top \Phi \mathbf{w}~&= (1 - \y) \frac{|\mathcal{D}|}{|\mathcal{D}_0|} \Psi^\top \mathbf{1} \\
     \mathbf{w}^*~&= (1 - \y) \frac{|\mathcal{D}|}{|\mathcal{D}_0|} (\Phi^\top \Phi)^{-1} \Psi^\top \mathbf{1}.
\end{align}

\hfill $\blacksquare$

\subsection{Theorem 3}

\begin{theorem}
If the deep SR is exact, such that $(1 - \y) \E_{s_0,a_0} \lb \psi^\pi(s_0,a_0) \rb = \E_{(s,a) \sim d^\pi} [\phi(s,a)]$, and the support of $d^\pi$ is contained in the dataset $\mathcal{D}$, then the optimizer $\mathbf{w}^*$ of \autoref{appendix:eqn:jw2}, as defined by \autoref{appendix:eqn:wstar1}, is the least squares estimator of $\sum_{(s,a) \in \mathcal{D}} \lp \mathbf{w}^\top \phi(s,a) - \frac{d^\pi(s,a)}{d^\mathcal{D}(s,a)} \rp^2$.
\end{theorem}

Where
\begin{equation} \label{appendix:eqn:jw2}
\min_{\mathbf{w}} J(\mathbf{w}) := \frac{1}{2|\mathcal{D}|} \sum_{(s,a) \in \mathcal{D}} \lb (\mathbf{w}^\top \phi(s,a))^2 \rb - (1 - \gamma) \frac{1}{|\mathcal{D}_0|}\sum_{s_0 \in \mathcal{D}_0, a_0} \pi(a_0|s_0) \mathbf{w}^\top \psi^\pi(s_0,a_0).
\end{equation}
and (as derived in \autoref{appendix:sec:wstar})
\begin{equation} \label{appendix:eqn:wstar1}
\mathbf{w}^* = (1 - \y) \frac{|\mathcal{D}|}{|\mathcal{D}_0|} (\Phi^\top \Phi)^{-1} \Psi^\top \mathbf{1}_{|\mathcal{D}_0||\mathcal{A}|}.
\end{equation}

\textit{Proof.}

Let: 
\begin{itemize}[nosep]
    \item $\Phi$ be a $|\mathcal{D}| \times F$ matrix where each row is the feature vector $\phi(s,a)$ with $F$ features.
    \item  $\Psi$ be a $|\mathcal{D}_0||\mathcal{A}| \times F$ matrix where each row is $\pi(a_0 |s_0) \psi^\pi(s_0,a_0)$, the SR weighted by its probability under the policy over all states $s_0$ in dataset of start states $\mathcal{D}_0$ and all actions. 
    \item $\mathbf{1}_x$ be a $x$ dimensional vector of all $1$.
    \item $d^\pi$ and $d^\mathcal{D}$ be diagonal $|\mathcal{D}| \times |\mathcal{D}|$ matrices where the diagonal entries contain $d^\pi(s,a)$ and $d^\mathcal{D}(s,a)$ respectively, for each state-action pair $(s,a)$ in the dataset $\mathcal{D}$.
\end{itemize}

First note the least squares estimator of $\sum_{(s,a) \in \mathcal{D}} \lp \mathbf{w}^\top \phi(s,a) - \frac{d^\pi(s,a)}{d^\mathcal{D}(s,a)} \rp^2$ is $(\Phi^\top \Phi)^{-1} \Phi^\top \frac{d^\pi}{d^\mathcal{D}} \mathbf{1}_{|\mathcal{D}|}$, where the division is element-wise.

By our assumption on the deep SR, we have that:
\begin{align}
(1 - \y) \E_{s_0,a_0} \lb \psi^\pi(s_0,a_0) \rb &= \E_{(s,a) \sim d^\pi} [\phi(s,a)] \\
&= \E_{(s,a) \sim d^\mathcal{D}} \lb \frac{d^\pi(s,a)}{d^\mathcal{D}(s,a)} \phi(s,a) \rb.
\end{align}
and therefore:
\begin{align}
    (1 - \y) \frac{1}{|\mathcal{D}_0|} \mathbf{1}_{|\mathcal{D}_0||\mathcal{A}|}^\top \Psi
    = \frac{1}{|\mathcal{D}|} \mathbf{1}_{|\mathcal{D}|}^\top \frac{d^\pi}{d^\mathcal{D}} \Phi.
\end{align}
It follows that we can simplify $\mathbf{w}^*$:
\begin{align}
    \mathbf{w}^*~&= (1 - \y) \frac{|\mathcal{D}|}{|\mathcal{D}_0|} (\Phi^\top \Phi)^{-1} \Psi^\top \mathbf{1}_{|\mathcal{D}_0||\mathcal{A}|} \\
    &= |\mathcal{D}| (\Phi^\top \Phi)^{-1} \lp (1 - \y) \frac{1}{|\mathcal{D}_0|} \mathbf{1}_{|\mathcal{D}_0||\mathcal{A}|}^\top \Psi \rp^\top \\
    &= |\mathcal{D}| (\Phi^\top \Phi)^{-1} \lp \frac{1}{|\mathcal{D}|} \mathbf{1}_{|\mathcal{D}|}^\top \frac{d^\pi}{d^\mathcal{D}} \Phi \rp^\top \\
    &= (\Phi^\top \Phi)^{-1} \Phi^\top \frac{d^\pi}{d^\mathcal{D}} \mathbf{1}_{|\mathcal{D}|}. 
\end{align}

\hfill $\blacksquare$

\subsection{Theorem 4}

\begin{theorem} 
Given the least squares estimator $\mathbf{w}_{\text{SR}}$ of $\sum_{(s,a) \in \mathcal{D}} \lp \mathbf{w}^\top \phi(s,a) - r(s,a) \rp^2$ and the optimizer $\mathbf{w}^*$ of \autoref{appendix:eqn:jw4}, as defined by \autoref{appendix:eqn:wstar}, then the traditional SR estimator $\frac{1}{|\mathcal{D}_0|} \sum_{s_0 \in \mathcal{D}_0} \mathbf{w}_\text{SR}^\top \psi^\pi(s_0,a_0)$ of $R(\pi)$ is identical to the SR-DICE estimator $\frac{1}{|\mathcal{D}|} \sum_{(s,a,r(s,a)) \in \mathcal{D}} \mathbf{w}^{*\top} \phi(s,a) r(s,a)$ of $R(\pi)$.
\end{theorem}

Where
\begin{equation} \label{appendix:eqn:jw4}
\min_{\mathbf{w}} J(\mathbf{w}) := \frac{1}{2|\mathcal{D}|} \sum_{(s,a) \in \mathcal{D}} \lb (\mathbf{w}^\top \phi(s,a))^2 \rb - (1 - \gamma) \frac{1}{|\mathcal{D}_0|}\sum_{s_0 \in \mathcal{D}_0, a_0} \pi(a_0|s_0) \mathbf{w}^\top \psi^\pi(s_0,a_0).
\end{equation}
and (as derived in \autoref{appendix:sec:wstar})
\begin{equation} \label{appendix:eqn:wstar}
    \mathbf{w}^* = (1 - \y) \frac{|\mathcal{D}|}{|\mathcal{S}_0|} (\Phi^\top \Phi)^{-1} \Psi^\top \mathbf{1}.
\end{equation}

\textit{Proof.}

Let:
\begin{itemize}[nosep]
    \item $\Phi$ be a $|\mathcal{D}| \times F$ matrix where each row is the feature vector $\phi(s,a)$ with $F$ features.
    \item  $\Psi$ be a $|\mathcal{D}_0||\mathcal{A}| \times F$ matrix where each row is $\pi(a_0 |s_0) \psi^\pi(s_0,a_0)$, the SR weighted by its probability under the policy over all states $s_0$ in dataset of start states $\mathcal{D}_0$ and all actions. 
    \item $\mathbf{1}$ be a $|\mathcal{D}_0||\mathcal{A}|$ dimensional vector of all $1$.
    \item $R$ be the $\mathcal{D}$ dimensional vector of each reward $r(s,a)$ in the dataset $\mathcal{D}$. 
\end{itemize}

First note the least squares solution for direct SR, where $\mathbf{w}_\text{SR}$ is optimized to reduce the mean squared error between $\mathbf{w}_\text{SR} \phi(s,a)$ and $r(s,a)$:
\begin{equation}
    \mathbf{w}_{\text{SR}} = (\Phi^\top \Phi)^{-1} \Phi^\top R.
\end{equation}

It follows that the direct SR solution to $R(\pi)$ is:
\begin{equation}
    (1 -\y) \frac{1}{|\mathcal{D}_0|} \mathbf{1}^\top \Psi \mathbf{w}_{\text{SR}} = (1 -\y) \frac{1}{|\mathcal{D}_0|} \mathbf{1}^\top \Psi (\Phi^\top \Phi)^{-1} \Phi^\top R.
\end{equation}

Now consider the SR-DICE solution to $R(\pi)$:
\begin{align}
    \frac{1}{|\mathcal{D}|} (\Phi \mathbf{w}^*)^\top R~&= \frac{1}{|\mathcal{D}|} \mathbf{w}^{*\top} \Phi^\top R \\
    &= \frac{1}{|\mathcal{D}|} \lp (1 - \y) \frac{|\mathcal{D}|}{|\mathcal{D}_0|} (\Phi^\top \Phi)^{-1} \Psi^\top \mathbf{1} \rp^\top \Phi^\top R \\ 
    &= (1 - \y)  \frac{1}{|\mathcal{D}_0|} \mathbf{1}^\top \Psi (\Phi^\top \Phi)^{-1} \Phi^\top R.
\end{align}

\hfill $\blacksquare$
\bigskip

\section{Additional Experiments}

In this section, we include additional experiments and visualizations, covering extra domains, additional ablation studies, run time experiments and additional behavior policies in the Atari domain. 

\subsection{Extra Continuous Domains}

\begin{figure}[ht]
\centering
\includegraphics[width=0.45\linewidth]{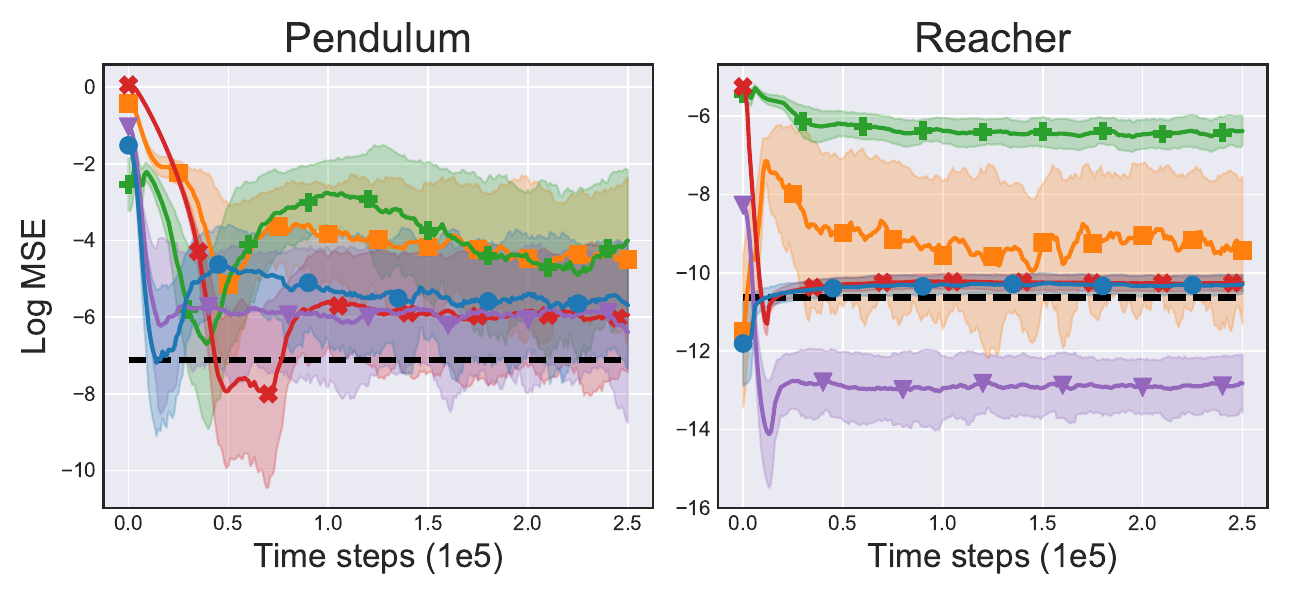}
\includegraphics[width=0.45\linewidth]{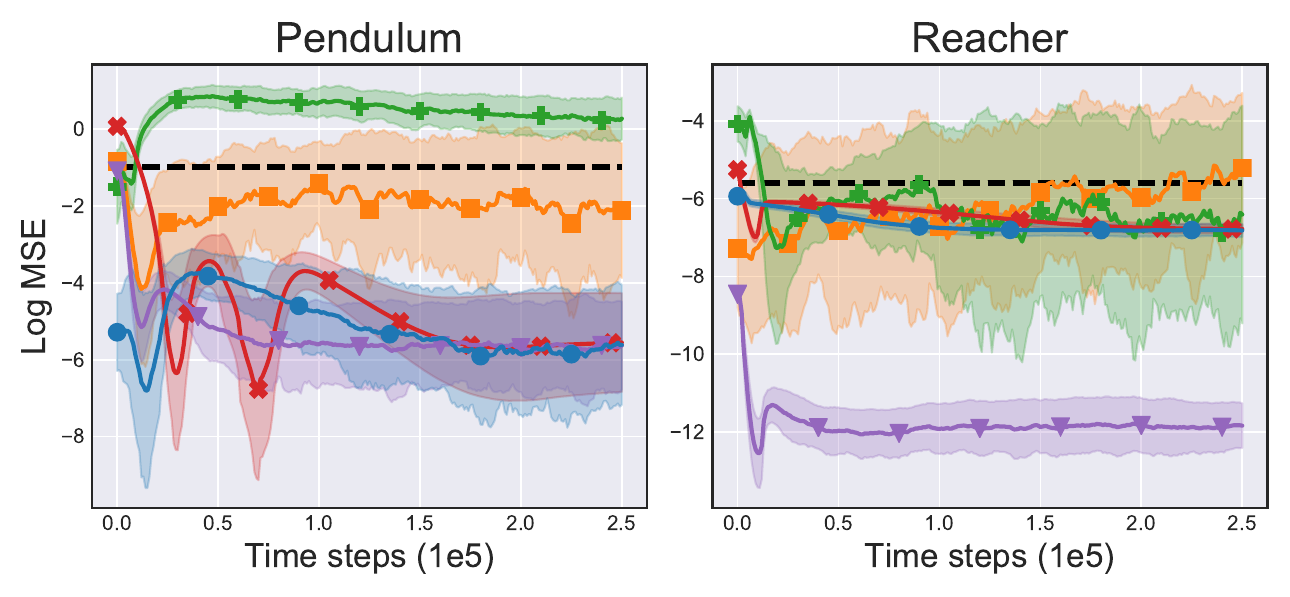}

\small{
\ccirc{31}{119}{180} SR-DICE \quad
\csquare{255}{127}{14} DualDICE \quad
\cplus{44}{160}{44} GradientDICE \quad
\ctimes{214}{39}{40} Deep SR \quad 
\ctriangle{148}{103}{189} Deep TD \quad
\cdashed Behavior $R(\pi_b)$
}
\vspace{-2mm}

\subfloat[\textit{Easy} setting (500k time steps and $\sigma_b=0.133$)]{\hspace{0.45\linewidth}}%
\subfloat[\textit{Hard} setting (50k time steps, $\sigma_b=0.2$, random actions with $p=0.2$)]{\hspace{0.45\linewidth}}%
\caption{Off-policy evaluation results for Pendulum and Reacher. The shaded area captures one standard deviation across 10 trials. Even on these easier environment, we find that SR-DICE outperforms the baseline MIS methods.} \label{fig:toy_results}
\vspace{-0.35cm}
\end{figure}

Although our focus is on high-dimensional domains, the environments, Pendulum and Reacher, have appeared in several related MIS papers \citep{nachum2019dualdice, zhang2020gendice}. Therefore, we have included results for these domains in \autoref{fig:toy_results}. All experimental settings match the experiments in the main body, and are described fully in Appendix \ref{appendix:section:details}.

\subsection{Representation Learning \& MIS}

SR-DICE relies a disentangled representation learning phase where an encoding $\phi$ is learned, followed by the deep successor representation $\psi^\pi$ which are used with a linear vector $\mathbf{w}$ to estimate the density ratios. In this section we perform some experiments which attempt to evaluate the importance of representation learning by comparing their influence on the baseline MIS methods. 

\begin{figure}[ht]
\centering
\includegraphics[width=0.225\linewidth]{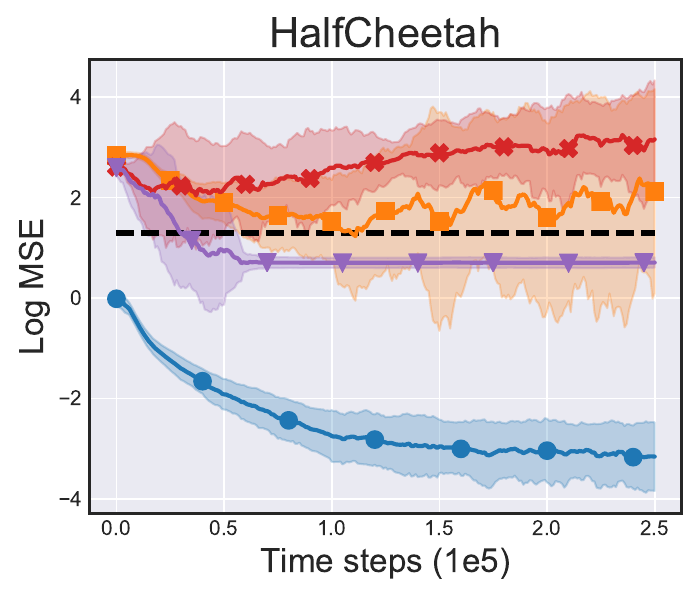}\hfill
\includegraphics[width=0.225\linewidth]{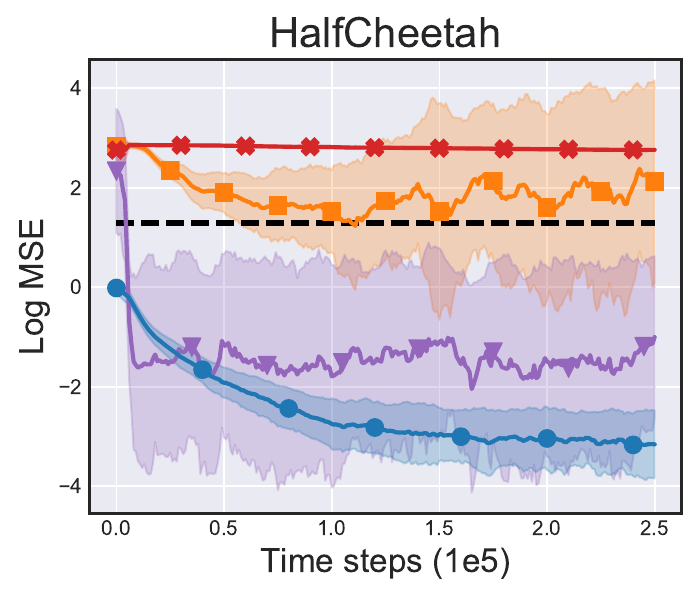}\hfill
\includegraphics[width=0.225\linewidth]{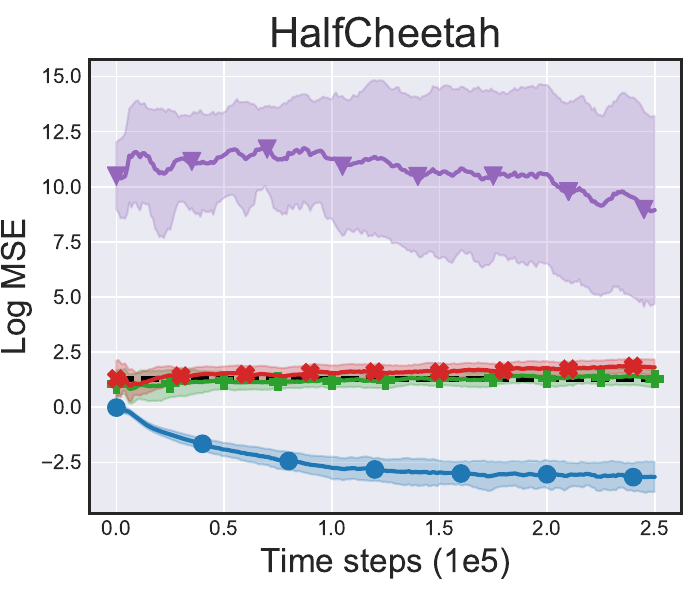}\hfill
\includegraphics[width=0.225\linewidth]{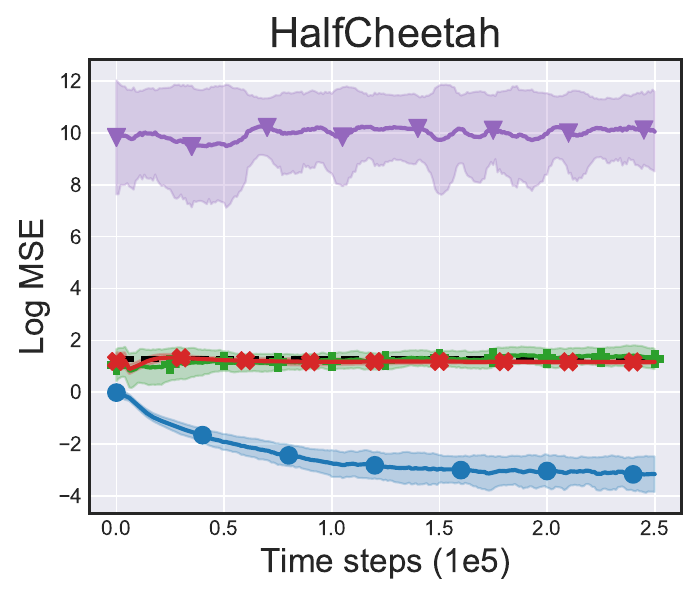}

\small{
\ccirc{31}{119}{180} SR-DICE \quad
\csquare{255}{127}{14} DualDICE \quad
\cplus{44}{160}{44} GradientDICE \quad
\ctimes{214}{39}{40} Encoding+X \quad 
\ctriangle{148}{103}{189} SR+X \quad
\cdashed Behavior $R(\pi_b)$
}
\vspace{-2mm}

\subfloat[DualDICE]{\hspace{0.225\linewidth}}\hfill
\subfloat[Linear DualDICE]{\hspace{0.225\linewidth}}\hfill
\subfloat[GradientDICE]{\hspace{0.225\linewidth}}\hfill
\subfloat[Linear GradientDICE]{\hspace{0.24\linewidth}}

\caption{Off-policy evaluation results on HalfCheetah examining the value of differing representations added to the baseline MIS methods. The experimental setting corresponds to the \textit{hard} setting from the main body. The shaded area captures one standard deviation across 10 trials. We see that using the SR $\psi^\pi$ as a representation improves the performance of DualDICE. On the other hand, GradientDICE performs much worse when using the SR, suggesting it cannot be used naively to improve MIS methods.}
\label{fig:phipsi}
\end{figure}

\textbf{Alternate representations.} We examine both DualDICE \citep{nachum2019dualdice} and GradientDICE~\citep{zhang2020gradientdice} under four settings where we pass the representations $\phi$ and $\psi^\pi$ to their networks, where both $\phi$ and $\psi^\pi$ are learned in identical fashion to SR-DICE.
\begin{enumerate}[nosep,leftmargin=2em,label=(\arabic*)]
    \item \makebox[6cm]{Input encoding $\phi$,\hfill} \makebox[2cm]{$f(\phi(s,a))$,\hfill} $w(\phi(s,a))$.
    \item \makebox[6cm]{Input SR $\psi^\pi$,\hfill} \makebox[2cm]{$f(\psi^\pi(s,a))$,\hfill} $w(\psi^\pi(s,a))$.
    \item \makebox[6cm]{Input encoding $\phi$, linear networks,\hfill} \makebox[2cm]{$f^\top \phi(s,a)$,\hfill} $w^\top \phi(s,a)$.
    \item \makebox[6cm]{Input SR $\psi^\pi$, linear networks,\hfill} \makebox[2cm]{$f^\top \psi^\pi(s,a)$,\hfill} $w^\top \psi^\pi(s,a)$.
\end{enumerate}
See Appendix \ref{appendix:section:baselines} for specific details on the baselines. We report the results in \autoref{fig:phipsi}. For GradientDICE, no benefit is provided by varying the representations, although using the encoding~$\phi$ matches the performance of vanilla GradientDICE regardless of the choice of network, providing some validation that $\phi$ is a reasonable encoding. Interestingly, for DualDICE, we see performance gains from using the SR $\psi^\pi$ as a representation: slightly as input, but significantly when used with linear networks. On the other hand, as GradientDICE performs much worse with the SR, it is clear that the SR cannot be used as a representation without some degree of forethought. 

\begin{figure}[ht]
    \centering
    \includegraphics[width=0.225\linewidth]{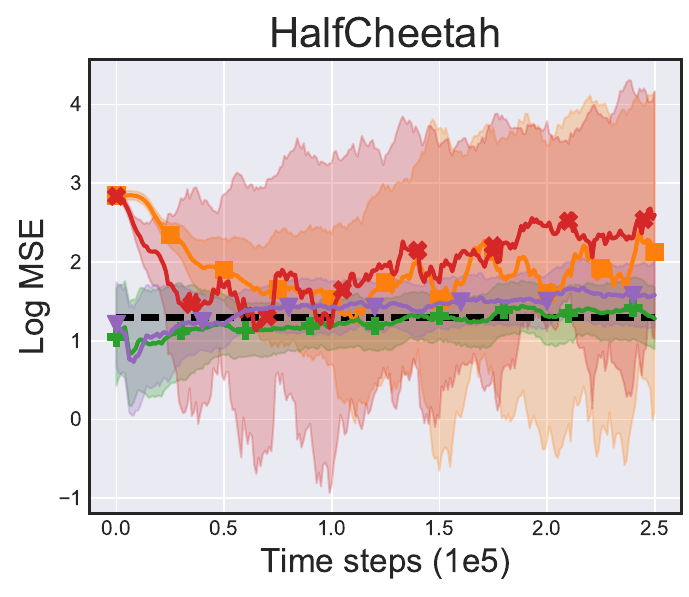}
    
\small{
\csquare{255}{127}{14} DualDICE \quad
\cplus{44}{160}{44} GradientDICE \quad
\ctimes{214}{39}{40} Big DualDICE \quad 
\ctriangle{148}{103}{189} Big GradientDICE \quad
\cdashed Behavior $R(\pi_b)$
}
    \caption{Off-policy evaluation results on HalfCheetah evaluating the performance benefits from larger network capacity on the baseline MIS methods. ``Big'' refers to the models with an additional hidden layer. The experimental setting corresponds to the \textit{hard} setting from the main body. The shaded area captures one standard deviation across 10 trials. We find that there is no clear performance benefit from increasing network capacity.}
    \label{fig:big}
\end{figure}

\textbf{Increased capacity.} As SR-DICE uses a linear function on top of a representation trained with the same capacity as the networks in DualDICE and GradientDICE, our next experiment examines if this additional capacity provides benefit to the baseline methods. To do, we expand each network in both baselines by adding an additional hidden layer. The results are reported in \autoref{fig:big}. We find there is a very slight decrease in performance when using the larger capacity networks. This suggests the performance gap from SR-DICE over the baseline methods has little to do with model~size.

\subsection{Toy Domains}

We additional test the MIS algorithms on a toy random-walk experiment with varying feature representations, based on a domain from \citep{tdc}. 

\textbf{Domain.} The domain is a simple 5-state MDP $(x_1,x_2,x_3,x_4,x_5)$ with two actions $(a_0,a_1)$, where action $a_0$ induces the transition $x_i \rightarrow x_{i-1}$ and action $a_1$ induces the transition $x_i \rightarrow x_{i+1}$, with the state $x_1$ looping to itself with action $a_0$ and $x_5$ looping to itself with action $a_5$. Episodes begin in the state $x_1$. 

\textbf{Target.} We evaluate policy $\pi$ which selects actions uniformly, i.e.\ $\pi(a_0|x_i) = \pi(a_1|x_i) = 0.5$ for all states $x_i$. Our dataset $\mathcal{D}$ contains all $10$ possible state-action pairs and is sampled uniformly. We use a discount factor of $\y=0.99$. Methods are evaluated on the average MSE between their estimate of $\frac{d^\pi}{d^\mathcal{D}}$ on all state-action pairs and the ground-truth value, which is calculated analytically. 

\textbf{Hyper-parameters.} Since we are mainly interested in a function approximation setting, each method uses a small neural network with two hidden layers of $32$, followed by tanh activation functions. All networks used stochastic gradient descent with a learning rate $\al$ tuned for each method out of $\{ 1, 0.5, 0.1, 0.05, 0.01, 0.001 \}$. This resulted in $\al=0.05$ for DualDICE, $\al=0.1$ for GradientDICE, and $\al=0.05$ for SR-DICE. Although there are a small number of possible data points, we use a batch size of $128$ to resemble the regular training procedure. As recommended by the authors we use $\lambda=1$ for GradientDICE~\citep{zhang2020gradientdice}, which was not tuned. For SR-DICE, we update the target network at every time step $\tau=1$, which was not tuned. 

Since there are only $10$ possible state-action pairs, we use the closed form solution for the vector $\mathbf{w}$. Additionally, we skip the state representation phase of SR-DICE, instead learning the SR $\psi^\pi$ over the given representation of each state, such that the encoding $\phi=x$. This allows us to test SR-DICE to a variety of representations rather than using a learned encoding. Consequently, with these choices, SR-DICE has no pre-training phase, and therefore, unlike every other graph in this paper, we report the results as the SR is trained, rather than as the vector $\mathbf{w}$ is trained.

\textbf{Features.} To test the robustness of each method we examine three versions of the toy domain, each using a different feature representation over the same 5-state MDP. These feature sets are again taken from \citep{tdc}.

\begin{itemize}[nosep,leftmargin=2em]
\item Tabular features: states are represented by a one-hot encoding, for example $x_2 = \lb 0, 1, 0, 0, 0 \rb$.
\item Inverted features: states are represented by the inverse of a one-hot encoding, for example $x_2 = \lb \frac{1}{2}, 0, \frac{1}{2}, \frac{1}{2}, \frac{1}{2} \rb$.
\item Dependent features: states are represented by 3 features which is not sufficient to cover all states exactly. In this case $x_1 = \lb 1, 0, 0 \rb$, $x_2 = [ \frac{1}{\sqrt{2}}, \frac{1}{\sqrt{2}}, 0 ]$, $x_3 = [ \frac{1}{\sqrt{3}}, \frac{1}{\sqrt{3}}, \frac{1}{\sqrt{3}} ]$, $x_4 = [ 0, \frac{1}{\sqrt{2}}, \frac{1}{\sqrt{2}} ]$, $x_5 = \lb 0, 0, 1 \rb$. Since our experiments use neural networks rather than linear functions, this representation is mainly meant to test SR-DICE, where we skip the state representation phase for SR-DICE and use the encoding $\phi=x$, limiting the representation of the SR. 
\end{itemize}

\begin{figure}[ht]
\centering
\includegraphics[width=0.675\linewidth]{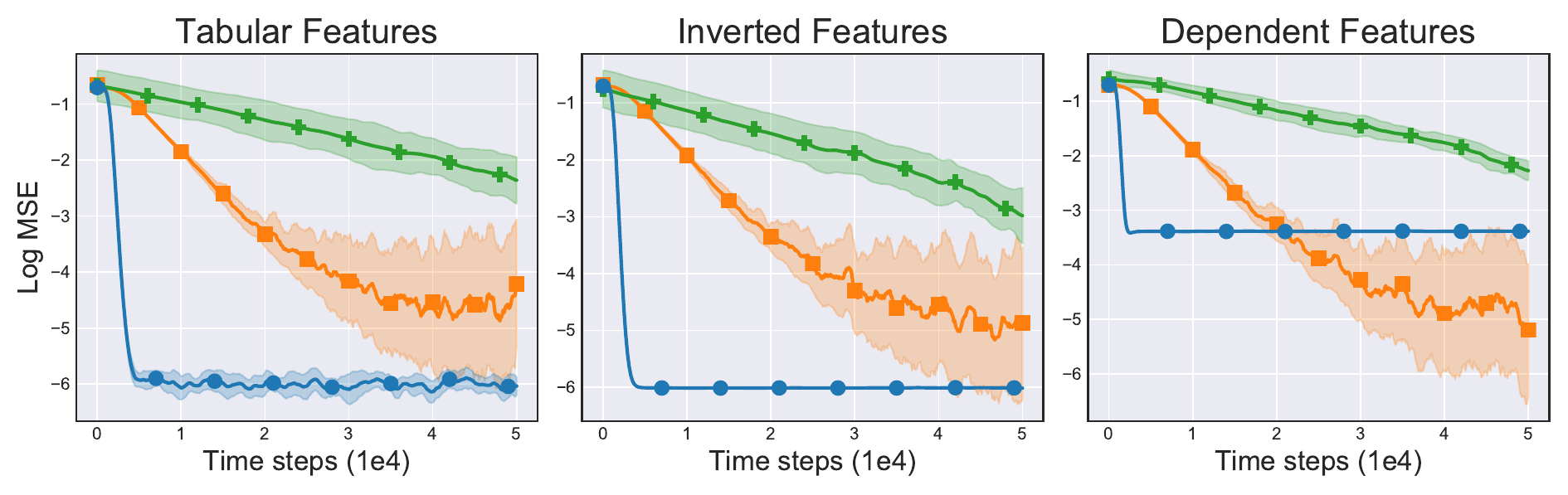}

\small{
\ccirc{31}{119}{180} SR-DICE \quad
\csquare{255}{127}{14} DualDICE \quad
\cplus{44}{160}{44} GradientDICE
}
\vspace{-2mm}

\subfloat[Tabular Features]{\hspace{0.225\linewidth}}%
\subfloat[Inverted Features]{\hspace{0.24\linewidth}}%
\subfloat[Dependent Features]{\hspace{0.225\linewidth}}%

\caption{Results measuring the log MSE between the estimated density ratio and the ground-truth on a simple 5-state MDP domain with three feature sets. The shaded area captures one standard deviation across 10 trials. Results are evaluated every $100$ time steps over $50$k time steps total.}
\label{fig:tabular}
\end{figure}

\textbf{Results.} We report the results in \autoref{fig:tabular}. We remark on several observations. SR-DICE learns significantly faster than the baseline methods, likely due to its use of temporal difference methods in the SR update, rather than using an update similar to residual learning, which is notoriously slow~\citep{baird1995residual, zhang2020deep}. GradientDICE appears to still be improving, although we limit training at $50$k time steps, which we feel is sufficient given the domain is deterministic and only has $5$ states. Notably, GradientDICE also uses a higher learning rate than SR-DICE and DualDICE. We also find the final performance of SR-DICE is much better than DualDICE and GradientDICE in the domains where the feature representation is not particularly destructive, highlighting the easier optimization of SR-DICE. In the case of the dependent features, we find DualDICE outperforms SR-DICE after sufficient updates. However, we remark that this concern could likely be resolved by learning the features and that SR-DICE still outperforms GradientDICE. Overall, we believe these results demonstrate that SR-DICE's strong empirical performance is consistent across simpler domains as well as the high-dimensional domains we examine in the main body. 

\subsection{Run Time Experiments} \label{appendix:section:runtime}

In this section, we evaluate the run time of each algorithm used in our experiments. Although SR-DICE relies on pre-training the deep successor representation before learning the density ratios, we find each marginalized importance sampling (MIS) method uses a similar amount of compute, due to the reduced cost of training $\mathbf{w}$ after the pre-training phase. 

We evaluate the run time on the HalfCheetah environment in MuJoCo \citep{mujoco} and OpenAI gym \citep{OpenAIGym}. As in the main set of experiments, each method is trained for $250$k time steps. Additionally, SR-DICE and Deep SR train the encoder-decoder for $30$k time steps and the deep successor representation for $100$k time steps before training $\mathbf{w}$. Run time is averaged over $3$ seeds. All time-based experiments are run on a single GeForce GTX 1080 GPU and a Intel Core i7-6700K CPU. Results are reported in \autoref{appendix:fig:run_time}. 

\begin{figure}[ht]
    \centering
    \includegraphics[width=0.5\linewidth]{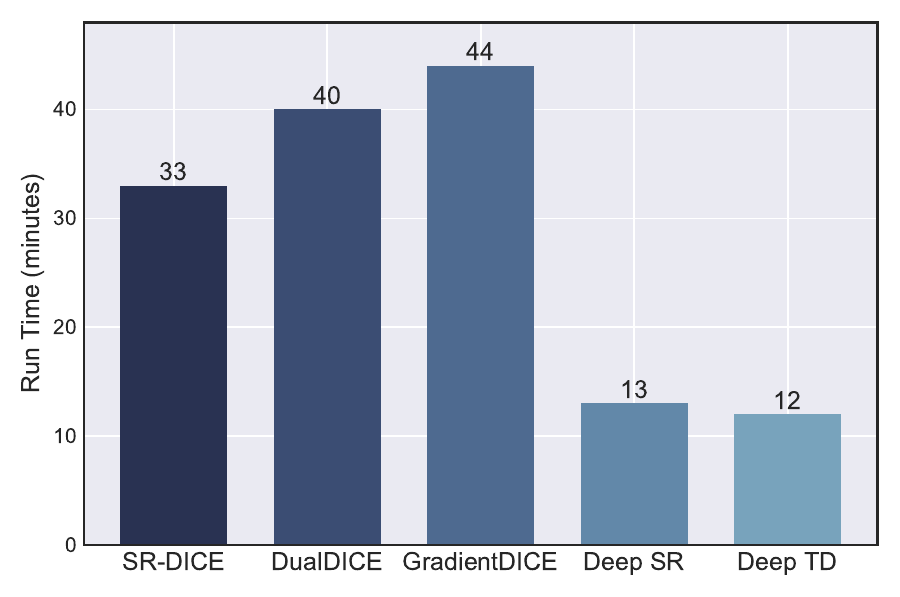}
    \caption{The average run time of each off-policy evaluation approach in minutes. Each experiment is run for $250$k time steps and is averaged over $3$ seeds. SR-DICE and Deep SR pre-train encoder-decoder for $30$k time steps and the deep successor representation $100$k time steps.}
    \label{appendix:fig:run_time}
\end{figure}

We find the MIS algorithms run in a comparable time, regardless of the pre-training step involved in SR-DICE. This can be explained as training $\mathbf{w}$ in SR-DICE involves significantly less compute than DualDICE and GradientDICE which update multiple networks. On the other hand, the deep reinforcement learning approaches run in about half the time of SR-DICE.

\subsection{Atari Experiments} \label{appendix:section:extra_atari}

To better evaluate the algorithms in the Atari domain, we run two additional experiments where we swap the behavior policy. We observe similar trends as the experiments in the main body of the paper. In both experiments we keep all other settings fixed. Notably, we continue to use the same target policy, corresponding to the greedy policy trained by Double DQN~\citep{DoubleDQN}, the same discount factor $\y=0.99$, and the same dataset size of $1$ million. 

\textbf{Increased noise.} In our first experiment, we attempt to increase the randomness of the behavior policy. As this can cause destructive behavior in the performance of the agent, we adopt an episode-dependent policy which selects between the noisy policy or the deterministic greedy policy at the beginning of each episode. This is motivated by the offline deep reinforcement learning experiments from \citep{fujimoto2019benchmarking}. As a result, we use an $\e$-greedy policy with $p=0.8$ and the deterministic greedy policy (the target policy) with $p=0.2$. $\e$ is set to $0.2$, rather than $0.1$ as in the experiments in the main body of the paper. Results are reported in \autoref{appendix:fig:atari_results_02}.

\begin{figure}[ht]
    \centering
    \includegraphics[width=\linewidth]{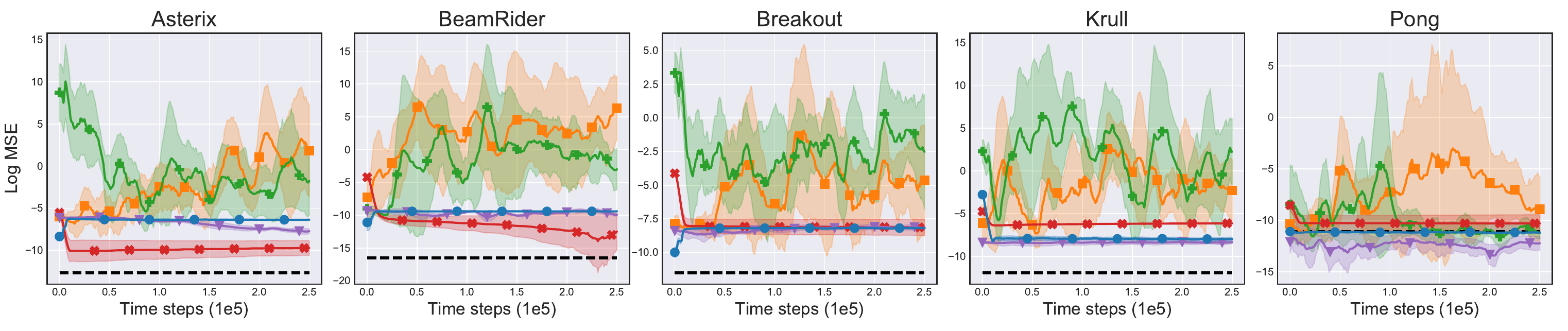}
\small{
\ccirc{31}{119}{180} SR-DICE \quad
\csquare{255}{127}{14} DualDICE \quad
\cplus{44}{160}{44} GradientDICE \quad
\ctimes{214}{39}{40} Deep SR \quad 
\ctriangle{148}{103}{189} Deep TD \quad
\cdashed Behavior $R(\pi_b)$
}
    \caption{We plot the log MSE for off-policy evaluation in the image-based Atari domain, using an episode-dependent noisy policy, where $\e=0.2$ with $p=0.8$ and $\e=0$ with $p=0.2$. This episode-dependent selection ensures sufficient state-coverage while using a stochastic policy. The shaded area captures one standard deviation across 3 trials. Markers are not placed at every point for visual clarity.}
    \label{appendix:fig:atari_results_02}
\end{figure}

We observe very similar trends to the original set of experiments. Again, we note DualDICE and GradientDICE perform very poorly, while SR-DICE, Deep SR, and Deep TD achieve a reasonable, but biased, performance. In this setting, we still find the behavior policy is the closest estimate of the true value of $R(\pi)$ .

\textbf{Separate behavior policy.} In this experiment, we use a behavior which is distinct from the target policy, rather than simply adding noise. This behavior policy is derived from an agent trained with prioritized experience replay and Double DQN \citep{PrioritizedExpReplay, fujimoto2020equivalence}. Again, we use a $\e$-greedy policy, with $\e=0.1$. We report the results in \autoref{appendix:fig:atari_results_new}.

\begin{figure}[ht]
    \centering
    \includegraphics[width=\linewidth]{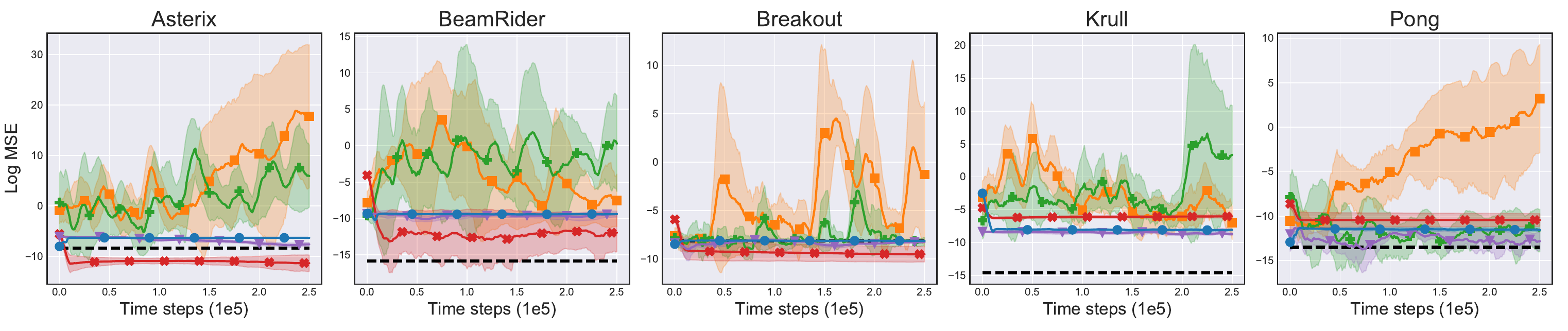}
\small{
\ccirc{31}{119}{180} SR-DICE \quad
\csquare{255}{127}{14} DualDICE \quad
\cplus{44}{160}{44} GradientDICE \quad
\ctimes{214}{39}{40} Deep SR \quad 
\ctriangle{148}{103}{189} Deep TD \quad
\cdashed Behavior $R(\pi_b)$
}    
    \caption{We plot the log MSE for off-policy evaluation in the image-based Atari domain, using a distinct behavior policy, trained by a separate algorithm, from the target policy. This experiment tests the ability to generalize to a more off-policy setting. The shaded area captures one standard deviation across 3 trials. Markers are not placed at every point for visual clarity.} \label{appendix:fig:atari_results_new}
\end{figure}

Again, we observe similar trends in performance. Notably, in the Asterix game, the performance of Deep SR surpasses the behavior policy, suggesting off-policy evaluation can outperform the na\"ive estimator in settings where the policy is sufficiently ``off-policy'' and distinct.

\section{SR-DICE Practical Details} \label{appendix:section:SRDICE}

\begin{algorithm}[t]
  \caption{SR-DICE} \label{algorithm:SRDICE}
\begin{algorithmic}%
	\STATE \textbf{Input:} dataset $\mathcal{D}$, target policy $\pi$, number of iterations $T_1$, $T_2$, $T_3$, mini-batch size $N$, target\_update\_rate. 
	\STATE \hrulefill
	\STATE {\color{blue} \# Train the encoder-decoder.}
	\FOR{$t=1$ {\bfseries to} $T_1$} %
	\STATE Sample mini-batch of $N$ transitions $(s, a, r, s')$ from $\mathcal{D}$.
	\STATE $\min_{\phi, D_{s'}, D_a, D_r} \lambda_{s'} (D_{s'}(\phi(s,a)) - s')^2$
	\STATE \qquad $+ \lambda_{a} (D_{a}(\phi(s,a)) - a)^2 + \lambda_{r} (D_{r}(\phi(s,a)) - r)^2$.
	\ENDFOR %
	\STATE \hrulefill
    \STATE {\color{blue} \# Train the successor representation.}
	\FOR{$t=1$ {\bfseries to} $T_2$} %
	\STATE Sample mini-batch of $N$ transitions $(s, a, r, s')$ from $\mathcal{D}$.
	\STATE Sample $a' \sim \pi(s')$.
	\STATE $\min_{\psi^\pi} (\phi(s,a) + \y \psi'(s',a') - \psi^\pi(s,a))^2$.
	\STATE If $t \text{ mod target\_update\_rate} = 0$: $\psi' \leftarrow \psi$.
	\ENDFOR %
	\STATE \hrulefill
	\STATE {\color{blue} \# Learn $\mathbf{w}$.}
	\FOR{$t=1$ {\bfseries to} $T_3$} %
	\STATE Sample mini-batch of $N$ transitions $(s, a, r, s')$ from $\mathcal{D}$. %
	\STATE Sample mini-batch of $N$ start states $s_0$ from $\mathcal{D}$.
	\STATE Sample $a_0 \sim \pi(s_0)$.
	\STATE $\min_{\mathbf{w}} \frac{1}{2} (\mathbf{w}^\top \phi(s,a))^2 - (1 - \y) \mathbf{w}^\top \psi^\pi(s_0,a_0)$.
	\ENDFOR %
\end{algorithmic}
\end{algorithm}

In this section, we cover some basic implementation-level details of SR-DICE. Note that code is provided for additional clarity. 

SR-DICE uses two parametric networks, an encoder-decoder network to learn the encoding $\phi$ and a deep successor representation network $\psi^\pi$. Additionally, SR-DICE uses the weights of a linear function $\mathbf{w}$. SR-DICE begins by pre-training the encoder-decoder network and the deep successor representation before applying updates to $\mathbf{w}$. 

\textbf{Encoder-Decoder.} This encoder-decoder network encodes $(s,a)$ to the feature vector $\phi(s,a)$, which is then decoded by several decoder heads. For the Atari domain, we choose to condition the feature vector only on states $\phi(s)$, as the reward is generally independent of the action selection. This change applies to both SR-DICE and Deep SR. Most design decisions are inspired by prior work \citep{machado2017eigenoption, machado2018count}.

For continuous control, given a mini-batch transition $(s,a,r,s')$, the encoder-decoder network is trained to map the state-action pair $(s,a)$ to the next state $s'$, the action $a$ and reward $r$. The resulting loss function is as follows:
\begin{equation}
    \min_{\phi, D_{s'}, D_a, D_r} \Loss(\phi, D) := \lambda_{s'} (D_{s'}(\phi(s,a)) - s')^2 + \lambda_{a} (D_{a}(\phi(s,a)) - a)^2 + \lambda_{r} (D_{r}(\phi(s,a)) - r)^2.
\end{equation}
We use $\lambda_{s'} = 1$, $\lambda_{a} = 1$ and $\lambda_{r} = 0.1$. 

For the Atari games, given a mini-batch transition $(s,a,r,s')$, the encoder-decoder network is trained to map the state $s$ to the next state $s'$ and reward $r$, while penalizing the size of $\phi(s)$. The resulting loss function is as follows:
\begin{equation}
    \min_{\phi, D_{s'}, D_r} \Loss(\phi, D) := \lambda_{s'} (D_{s'}(\phi(s)) - s')^2 + \lambda_{r} (D_{r}(\phi(s)) - r)^2 + \lambda_\phi \phi(s)^2.
\end{equation}
We use $\lambda_{s'} = 1$, $\lambda_{r} = 0.1$ and $\lambda_\phi = 0.1$.

\textbf{Deep Successor Representation.} The deep successor representation $\psi^\pi$ is trained to estimate the accumulation of $\phi$. The training procedure resembles standard deep reinforcement learning algorithms. Given a mini-batch of transitions $(s,a,r,s')$ the network is trained to minimize the following loss:
\begin{equation}
\min_{\psi^\pi} \Loss(\psi^\pi) := (\phi(s,a) + \y \psi'(s',a') - \psi^\pi(s,a))^2,
\end{equation}
where $\psi'$ is the target network. A target network is a frozen network used to provide stability~\citep{DQN, kulkarni2016deep} in the learning target. The target network is updated to the current network $\psi' \leftarrow \psi^\pi$ after a fixed number of time steps, or updated with slowly at each time step $\psi' \leftarrow \tau \psi^\pi + (1 - \tau) \psi^\pi$ \citep{DDPG}. 

\textbf{Marginalized Importance Sampling Weights.} As described in the main body, we learn $\mathbf{w}$ by optimizing the following objective:
\begin{equation}
    \min_{\mathbf{w}} J(\mathbf{w}) := \frac{1}{2} \E_{(s,a) \sim d^\mathcal{D}} \lb (\mathbf{w}^\top \phi(s,a))^2 \rb - (1 - \gamma) \E_{s_0,a_0 \sim \pi} \lb \mathbf{w}^\top \psi^\pi(s_0,a_0) \rb.
\end{equation}
This is achieved by sampling state-action pairs uniformly from the dataset $\mathcal{D}$, alongside a mini-batch of start states $s_0$, which are recorded at the beginning of each episode during data collection. 

We summarize the learning procedure of SR-DICE in \autoref{algorithm:SRDICE}.

\section{Baselines}\label{appendix:section:baselines}

In this section, we cover some of the practical details of each of the baseline methods. 

\subsection{DualDICE}

Dual stationary DIstribution Correction Estimation (DualDICE) \citep{nachum2019dualdice} uses two networks $f$ and $w$. The general optimization problem is defined as follows:
\begin{equation}
\begin{aligned}
\min_f \max_w J(f,w):=~&\E_{(s,a) \sim d^\mathcal{D},a' \sim \pi, s'} \lb w(s,a) (f(s,a) - \y f(s',a')) - 0.5 w(s,a)^2 \rb \\
&- (1 - \gamma) \E_{s_0,a_0}[f(s_0,a_0)].
\end{aligned}
\end{equation}
In practice this corresponds to alternating single gradient updates to $f$ and $w$. The authors suggest possible alternative functions to the convex function $0.5 w(s,a)^2$ such as $\frac{2}{3} |w(s,a)|^\frac{3}{2}$, however in practice we found $0.5 w(s,a)^2$ performed the best. 

\subsection{GradientDICE}

Gradient stationary DIstribution Correction Estimation (GradientDICE) \citep{zhang2020gradientdice} uses two networks $f$ and $w$, and a scalar $u$. The general optimization problem is defined as follows:
\begin{equation}
\begin{aligned}
\min_w \max_{f, u} J(w,u,f):=~& (1 - \gamma) \E_{s_0,a_0}[f(s_0,a_0)] + \y \E_{(s,a) \sim d^\mathcal{D},a' \sim \pi, s'}[w(s,a) f(s',a')] \\
&- \E_{(s,a) \sim d^\mathcal{D}}[w(s,a) f(s,a)] + \lambda  \lp \E_{(s,a) \sim d^\mathcal{D}}[u w(s,a) - u] - 0.5 u^2 \rp.
\end{aligned}
\end{equation}
Similarly to DualDICE, in practice this involves alternating single gradient updates to $w$, $u$ and $f$. As suggested by the authors we use $\lambda=1$. 

\subsection{Deep SR}

Our Deep SR baseline is a policy evaluation version of deep successor representation \citep{kulkarni2016deep}. The encoder-decoder network and deep successor representation are trained in exactly the same manner as SR-DICE (see Section \ref{appendix:section:SRDICE}). Then, rather than train $\mathbf{w}$ to learn the marginalized importance sampling ratios, $\mathbf{w}$ is trained to recover the original reward function. Given a mini-batch of transitions $(s,a,r,s')$, the following loss is applied: 
\begin{equation}
    \min_\mathbf{w} \Loss(\mathbf{w}) := (r - \mathbf{w}^\top \phi(s,a))^2.
\end{equation}

\subsection{Deep TD}

Deep TD, short for deep temporal-difference learning, takes the standard deep reinforcement learning methodology, akin to DQN \citep{DQN}, and applies it to off-policy evaluation. Given a mini-batch of transitions $(s,a,r,s')$ the Q-network is updated by the following loss:
\begin{equation}
\min_{Q^\pi} \Loss(Q^\pi): = (r + \y Q'(s',a') - Q^\pi(s,a))^2,
\end{equation}
where $a'$ is sampled from the target policy $\pi(\cdot|s')$. Similarly, to training the deep successor representation, $Q'$ is a frozen target network which is updated to the current network after a fixed number of time steps, or incrementally at every time step. 

\section{Experimental Details} \label{appendix:section:details}

All networks are trained with PyTorch (version 1.4.0) \citep{paszke2019pytorch}. Any unspecified hyper-parameter uses the PyTorch default setting. 

\textbf{Evaluation.} The marginalized importance sampling methods are measured by the average weighted reward from transitions sampled from a replay buffer $\frac{1}{N} \sum_{(s,a,r)} w(s,a) r(s,a)$, with $N=10$k, while the deep RL methods use $\frac{(1 - \y)}{M} \sum_{s_0} Q(s_0, \pi(a_0))$, where $M$ is the number of episodes. Each OPE method is trained on data collected by some behavioral policy $\pi_b$. We estimate the ``true'' normalized average discounted reward of the target and behavior policies from $100$ roll-outs in the environment.

\subsection{Continuous Action Environments} \label{subsec:contaction}

Our agents are evaluated via tasks interfaced through OpenAI gym (version 0.17.2) \citep{OpenAIGym}, which mainly rely on the MuJoCo simulator (mujoco-py version 1.50.1.68) \citep{mujoco}. We provide a description of each environment in \autoref{table:cont_environments}.

\begin{table}[ht]%
\centering
\caption{Continuous action environment descriptions.} \label{table:cont_environments}
\begin{center}
\begin{tabular}{lccccc}
\toprule
Environment & State dim. & Action dim. & Episode Horizon & Task description \\
\midrule
Pendulum-v0 & $3$ & $1$ & $200$ & Balance a pendulum. \\
Reacher-v2 & $11$ & $2$ & $50$ & Move end effector to goal. \\
HalfCheetah-v3 & $17$ & $6$ & $1000$ & Locomotion. \\
Hopper-v3 & $11$ & $3$ & $1000$ & Locomotion. \\
Walker2d-v3 & $17$ & $6$ & $1000$ & Locomotion. \\
Ant-v3 & $111$ & $8$ & $1000$ & Locomotion. \\
Humanoid-v3 & $376$ & $17$ & $1000$ & Locomotion. \\
\bottomrule
\end{tabular}
\end{center}
\end{table}

\textbf{Experiments.} Our experiments are framed as off-policy evaluation tasks in which agents aim to evaluate $R(\pi) = \E_{(s,a)\sim d^\pi,r}[r(s,a)]$ for some target policy $\pi$. In each of our experiments, $\pi$ corresponds to a noisy version of a policy trained by a TD3 agent \citep{fujimoto2018addressing}, a commonly used deep reinforcement learning algorithm. Denote $\pi_d$, the deterministic policy trained by TD3 using the author's GitHub \url{https://github.com/sfujim/TD3}. The target policy is defined as: $\pi + \N(0, \sigma^2)$, where $\sigma=0.1$. The off-policy evaluation algorithms are trained on a dataset generated by a single behavior policy $\pi_b$. The experiments are done with two settings \textit{easy} and \textit{hard} which vary the behavior policy and the size of the dataset. All other settings are kept fixed. For the \textit{easy} setting the behavior policy is defined as:
\begin{equation}
    \pi_b = \pi_d +  \N(0, \sigma_b^2), \sigma_b=0.133,
\end{equation}
and $500$k time steps are collected (approximately $500$ trajectories for most tasks). The \textit{easy} setting is roughly based on the experimental setting from \citet{zhang2020gendice}. For the \textit{hard} setting the behavior policy adds an increased noise and selects random actions with $p=0.2$:
\begin{equation}
    \pi_b = 
    \begin{cases}
    \pi_d + \N(0, \sigma_b^2), \sigma_b=0.2 & p=0.8, \\
    \text{Uniform random action} & p=0.2,
    \end{cases}
\end{equation}
and only $50$k time steps are collected (approximately $50$ trajectories for most tasks). For Pendulum-v0 and Humanoid-v3, the range of actions is $[-2,2]$ and $[-0.4,0.4]$ respectively, rather than $[-1,1]$, so we scale the size of the noise added to actions accordingly.  
We set the discount factor to $\y=0.99$. All continuous action experiments are over $10$ seeds.

\textbf{Pre-training.} Both SR-DICE and Deep SR rely on pre-training the encoder-decoder and deep successor representation $\psi$. These networks were trained for $30$k and $100$k time steps respectively. As noted in Section \ref{appendix:section:runtime}, even when including this pre-training step, both algorithm have a lower running time than DualDICE and GradientDICE. 

\textbf{Architecture.} For fair comparison, we use the same architecture for all algorithms except for DualDICE. This a fully connected neural network with 2 hidden layers of $256$ and ReLU activation functions. This architecture was based on the network defined in the TD3 GitHub and was not tuned. For DualDICE, we found tanh activation functions improved stability over ReLU. 

For SR-DICE and SR-Direct we use a separate architecture for the encoder-decoder network. The encoder is a network with a single hidden layer of $256$, making each $\phi(s,a)$ a feature vector of $256$. There are three decoders for reward, action, and next state, respectively. For the action decoder and next state decoder we use a network with one hidden layer of $256$. The reward decoder is a linear function of the encoding, without biases. All hidden layers are followed by ReLU activation functions.

\textbf{Network hyper-parameters.} All networks are trained with the Adam optimizer \citep{adam}. We use a learning rate of $3$e$-4$, again based on TD3 for all networks except for GradientDICE, which we found required careful tuning to achieve a reasonable performance. For GradientDICE we found a learning rate of $1$e$-5$ for $f$ and $w$, and $1$e$-2$ for $u$ achieved the highest performance. For DualDICE we chose the best performing learning rate out of $\{1 \text{e} -2, 1 \text{e} -3, 3 \text{e} -4, 5 \text{e} -5, 1 \text{e} -5\}$. 
SR-DICE, Deep SR, and Deep TD were not tuned and use default hyper-parameters from deep RL algorithms. For training $\psi^\pi$ and $Q^\pi$ for the deep reinforcement learning aspects of SR-DICE, Deep SR, and Deep TD we use a mini-batch size of $256$ and update the target networks using $\tau=0.005$, again based on TD3. For all MIS methods, we use a mini-batch size of $2048$ as described by \citep{nachum2019dualdice}. We found SR-DICE and DualDICE succeeded with lower mini-batch sizes but did not test this in detail. All hyper-parameters are described in~\autoref{table:cont_hyperparameters}. 

\begin{table}[ht]\setlength{\tabcolsep}{5.8pt}
\centering
\caption{Continuous action environment training hyper-parameters.} \label{table:cont_hyperparameters}
\vspace{-4mm}
\begin{center}
\begin{tabular}{lccccc}
\toprule
Hyper-parameter & SR-DICE & DualDICE & GradientDICE & Deep SR & Deep TD \\
\midrule
Optimizer & Adam & Adam & Adam & Adam & Adam \\
$\psi^\pi$, $Q^\pi$ Learning rate & $3 \text{e} -4$ & - & - & $3 \text{e} -4$ & $3 \text{e} -4$ \\
$\mathbf{w}$ Learning rate & $3 \text{e} -4$ & - & - & $3 \text{e} -4$ & - \\
$f$ Learning rate & - & $5 \text{e} -5$  & $1 \text{e} -5$  & - & - \\
$w$ Learning rate & - & $5 \text{e} -5$  & $1 \text{e} -5$  & - & - \\
$u$ Learning rate & - & - & $1 \text{e} -2$ & - & - \\
$\psi^\pi$, $Q^\pi$ Mini-batch size & $256$ & - & - & $256$ & $256$ \\
$\mathbf{w}$, $f$, $w$, $u$, Mini-batch size & $2048$ & $2048$ & $2048$ & $2048$ & - \\
$\psi^\pi$, $Q^\pi$ Target update rate & $0.005$ & - & - & $0.005$ & $0.005$ \\
\bottomrule
\end{tabular}
\end{center}
\end{table}

\textbf{Visualizations.} We graph the log MSE between the estimate of $R(\pi)$ and the true $R(\pi)$, where the log MSE is computed as $\log{0.5 (X - R(\pi))^2}$. We smooth the learning curves over a uniform window of $10$. Agents were evaluated every $1$k time steps and performance is measured over $250$k time steps total. Markers are displayed every $25$k time steps with offset for visual clarity.

\textbf{Randomized reward experiments.} The randomized reward experiment uses the MIS ratios collected from the \textit{hard} setting from the continuous action environments. $1000$ reward functions were generated by a randomly initialized neural network. The neural network architecture is two hidden layers of $256$ with ReLU activation functions after each hidden layer and a sigmoid activation function after the final layer. Weights were sampled from the normal distribution and biases were set to $0$. The ground truth value was estimated from $100$ on-policy trajectories. If the ground truth value was less than $0.1$ or greater than $0.9$, (where the range of possible values is $[0,1]$), the reward function was considered redundant and removed. To reduce variance across reward functions, we normalize both the estimated $R(\pi)$ and ground truth $R(\pi)$ by dividing by the average reward for all state-action pairs in the dataset.

\subsection{Atari}

We interface with Atari by OpenAI gym (version 0.17.2) \citep{OpenAIGym}, all agents use the NoFrameskip-v0 environments that include sticky actions with $p=0.25$ \citep{machado2018revisiting}. 

\textbf{Pre-processing.} We use standard pre-processing steps based on \citet{machado2018revisiting} and  \citet{castro2018dopamine}. 
We base our description on \citep{fujimoto2019benchmarking}, which our code is closely based on. 
We define the following:
\begin{itemize}[nosep, leftmargin=*]
    \item Frame: output from the Arcade Learning Environment.
    \item State: conventional notion of a state in a MDP.
    \item Input: input to the network. 
\end{itemize}

The standard pre-processing steps are as follows:
\begin{itemize}[nosep, leftmargin=*]
    \item Frame: gray-scaled and reduced to $84 \times 84$ pixels, tensor with shape $(1,84,84)$.
    \item State: the maximum pixel value over the $2$ most recent frames, tensor with shape $(1,84,84)$.
    \item Input: concatenation over the previous $4$ states, tensor with shape $(4,84,84)$.
\end{itemize}
The notion of time steps is applied to states, rather than frames, and functionally, the concept of frames can be abstracted away once pre-processing has been applied to the environment. 

The agent receives a state every $4$th frame and selects one action, which is repeated for the following $4$ frames. If the environment terminates within these $4$ frames, the state received will be the last $2$ frames before termination. For the first $3$ time steps of an episode, the input, which considers the previous $4$ states, sets the non-existent states to all $0$s. An episode terminates after the game itself terminates, corresponding to multiple lives lost (which itself is game-dependent), or after $27$k time steps ($108$k frames or $30$ minutes in real time). Rewards are clipped to be within a range of $[-1,1]$.

Sticky actions are applied to the environment \citep{machado2018revisiting}, where the action $a_t$ taken at time step $t$, is set to the previously taken action $a_{t-1}$ with $p=0.25$, regardless of the action selected by the agent. Note this replacement is abstracted away from the agent and dataset. In other words, if the agent selects action $a$ at state $s$, the transition stored will contain $(s,a)$, regardless if $a$ is replaced by the previously taken action.

\textbf{Experiments.} For the main experiments we use a behavior and target policy derived from a Double DQN agent \citep{DoubleDQN}, a commonly used deep reinforcement learning algorithm. The behavior policy is an $\e$-greedy policy with $\e=0.1$ and the target policy is the greedy policy (i.e. $\e=0$). 
In Section \ref{appendix:section:extra_atari} we perform two additional experiments with a different behavior policy. Otherwise, all hyper-parameters are fixed across experiments. For each, the dataset contains $1$~million transitions and uses a discount factor of $\y=0.99$. Each experiment is evaluated over~$3$~seeds. 

\textbf{Pre-training.} Both SR-DICE and Deep SR rely on pre-training the encoder-decoder and deep successor representation $\psi$. Similar to the continuous action tasks, these networks were trained for $30$k and $100$k time steps respectively. 

\textbf{Architecture.} We use the same architecture as most value-based deep reinforcement learning algorithms for Atari, e.g.~\citep{DQN, DoubleDQN, PrioritizedExpReplay}. This architecture is used for all networks, other than the encoder-decoder network, for fair comparison and was not tuned in any way. 

The network has a 3-layer convolutional neural network (CNN) followed by a fully connected network with a single hidden layer. As mentioned in pre-processing, the input to the network is a tensor with shape $(4,84,84)$. The first layer of the CNN has a kernel depth of $32$ of size $8 \times 8$ and a stride of $4$. The second layer has a kernel depth of $32$ of size $4 \times 4$ and a stride of $2$. The third layer has a kernel depth of $64$ of size $3 \times 3$ and a stride of $1$. The output of the CNN is flattened to a vector of $3136$ before being passed to the fully connected network. The fully connected network has a single hidden layer of $512$. Each layer, other than the output layer, is followed by a ReLU activation function. The final layer of the network outputs $|\mathcal{A}|$ values where $|\mathcal{A}|$ is the number of actions. 

The encoder-decoder used by SR-DICE and SR-Direct has a slightly different architecture. The encoder is identical to the aforementioned architecture, except the final layer outputs the feature vector $\phi(s)$ with $256$ dimensions and is followed by a ReLU activation function. The next state decoder uses a single fully connected layer which transforms the vector of $256$ to $3136$ and then is passed through three transposed convolutional layers each mirroring the CNN. Hence, the first layer has a kernel depth of $64$, kernel size of $3 \times 3$ and a stride of $1$. The second layer has a kernel depth of $32$, kernel size of $4 \times 4$ and a stride of $2$. The final layer has a kernel depth of $32$, kernel size of $8 \times 8$ and a stride of $4$. This maps to a $(1,84,84)$ tensor. All layers other than the final layer are followed by ReLU activation functions. Although the input uses a history of the four previous states, as mentioned in the pre-processing section, we only reconstruct the succeeding state without history. We do this because there is overlap in the history of the current input and the input corresponding to the next time step. The reward decoder is a linear function without biases. 

\begin{table}[ht]\setlength{\tabcolsep}{5.5pt}
\centering
\caption{Training hyper-parameters for the Atari domain.} \label{appendix:table:atari_hyper}
\vspace{-4mm}
\begin{center}
\begin{tabular}{lccccc}
\toprule
Hyper-parameter & SR-DICE & DualDICE & GradientDICE & Deep SR & Deep TD \\
\midrule
Optimizer & Adam & Adam & Adam & Adam & Adam \\
$\psi^\pi$, $Q^\pi$ Learning rate & $6.25 \text{e} -5$ & - & - & $6.25 \text{e} -5$ & $6.25 \text{e} -5$ \\
$\psi^\pi$, $Q^\pi$, $f$, $w$ Adam $\e$ & $1.5$e$-4$ & $1.5$e$-4$ & $1.5$e$-4$ & $1.5$e$-4$ & $1.5$e$-4$ \\
$\mathbf{w}$, $u$ Adam $\e$ & $1$e$-8$ & - & $1$e$-8$ & $1$e$-8$ & - \\
$\mathbf{w}$ Learning rate & $3 \text{e} -4$ & - & - & $3 \text{e} -4$ & - \\
$f$ Learning rate & - & $6.25 \text{e} -5$  & $6.25 \text{e} -5$  & - & - \\
$w$ Learning rate & - & $6.25 \text{e} -5$  & $6.25 \text{e} -5$  & - & - \\
$u$ Learning rate & - & - & $1 \text{e} -3$ & - & - \\
$\psi^\pi$, $Q^\pi$ Mini-batch size & $32$ & - & - & $32$ & $32$ \\
$\mathbf{w}$, $f$, $w$, $u$, Mini-batch size & $32$ & $32$ & $32$ & $32$ & - \\
$\psi^\pi$, $Q^\pi$ Target update rate & $8$k & - & - & $8$k & $8$k \\
\bottomrule
\end{tabular}
\end{center}
\end{table}

\textbf{Network hyper-parameters.} Our hyper-parameter choices are based on standard hyper-parameters based largely on \citep{castro2018dopamine}. All networks are trained with the Adam optimizer \citep{adam}. We use a learning rate of $6.25$e$-5$. Although not traditionally though of has a hyper-parameter, in accordance to prior work, we modify $\e$ used by Adam to be $1.5$e$-4$. For $\mathbf{w}$ we use a learning rate of $3$e$-4$ with the default setting of $\e=1$e$-8$. For $u$ we use $1$e$-3$. We use a mini-batch size of $32$ for all networks. SR-DICE, Deep SR, and Deep TD update the target network every $8$k time steps. All hyper-parameters are described in \autoref{appendix:table:atari_hyper}.

\textbf{Visualizations.} We use identical visualizations to the continuous action environments. Graphs display the log MSE between the estimate of $R(\pi)$ and the true $R(\pi)$ of the target policy, where the log MSE is computed as $\log{0.5 (X - R(\pi))^2}$. We smooth the learning curves over a uniform window of $10$. Agents were evaluated every $1$k time steps and performance is measured over $250$k time steps total. Markers are displayed every $25$k time steps with offset for visual clarity.

\end{document}